\documentclass{article}

\usepackage{microtype}
\usepackage{graphicx}
\usepackage{booktabs} 

\usepackage{hyperref}

\usepackage[accepted]{icml2024}

\usepackage{amsmath}
\usepackage{amssymb}
\usepackage{mathtools}
\usepackage{amsthm}

\usepackage[capitalize,noabbrev]{cleveref}

\theoremstyle{plain}

\theoremstyle{definition}

\theoremstyle{remark}

\usepackage{url}
\usepackage{kotex}
\usepackage{multirow}
\usepackage{arydshln}
\usepackage{caption}
\usepackage{wrapfig}
\usepackage{algorithmic}
\usepackage{algorithm}
\usepackage{subcaption}

\usepackage[textsize=tiny]{todonotes}

\begin{document}

\twocolumn[
\icmltitle{PruNeRF: Segment-Centric Dataset Pruning via 3D Spatial Consistency} 



\icmlsetsymbol{equal}{*}

\begin{icmlauthorlist}
\icmlauthor{Yeonsung Jung}{kaist}
\icmlauthor{Heecheol Yun}{kaist}
\icmlauthor{Joonhyung Park}{kaist}
\icmlauthor{Jin-Hwa Kim}{naver,snu}
\icmlauthor{Eunho Yang}{kaist,aitrics}
\end{icmlauthorlist}

\icmlaffiliation{kaist}{Graduate School of AI, Korea Advanced Institute of Science and Technology (KAIST), Republic of Korea}
\icmlaffiliation{aitrics}{AITRICS, Republic of Korea}
\icmlaffiliation{naver}{NAVER AI Lab, Republic of Korea}
\icmlaffiliation{snu}{AI Institute of Seoul National University, Republic of Korea}

\icmlcorrespondingauthor{Yeonsung Jung}{ys.jung@kaist.ac.kr}
\icmlcorrespondingauthor{Jin-Hwa Kim}{j1nhwa.kim@navercorp.com}
\icmlkeywords{Machine Learning, ICML}

\vskip 0.3in
]



\printAffiliationsAndNotice{}  

\begin{abstract}
Neural Radiance Fields (NeRF) have shown remarkable performance in learning 3D scenes. However, NeRF exhibits vulnerability when confronted with distractors in the training images -- unexpected objects are present only within specific views, such as moving entities like pedestrians or birds. Excluding distractors during dataset construction is a straightforward solution, but without prior knowledge of their types and quantities, it becomes prohibitively expensive. In this paper, we propose PruNeRF, a segment-centric dataset pruning framework via 3D spatial consistency, that effectively identifies and prunes the distractors. We first examine existing metrics for measuring pixel-wise distraction and introduce Influence Functions for more accurate measurements. Then, we assess 3D spatial consistency using a depth-based reprojection technique to obtain 3D-aware distraction. Furthermore, we incorporate segmentation for pixel-to-segment refinement, enabling more precise identification. Our experiments on benchmark datasets demonstrate that PruNeRF consistently outperforms state-of-the-art methods in robustness against distractors.
\end{abstract}

\begin{figure*}[t]
    \begin{center}
    \includegraphics[width=0.85\textwidth]{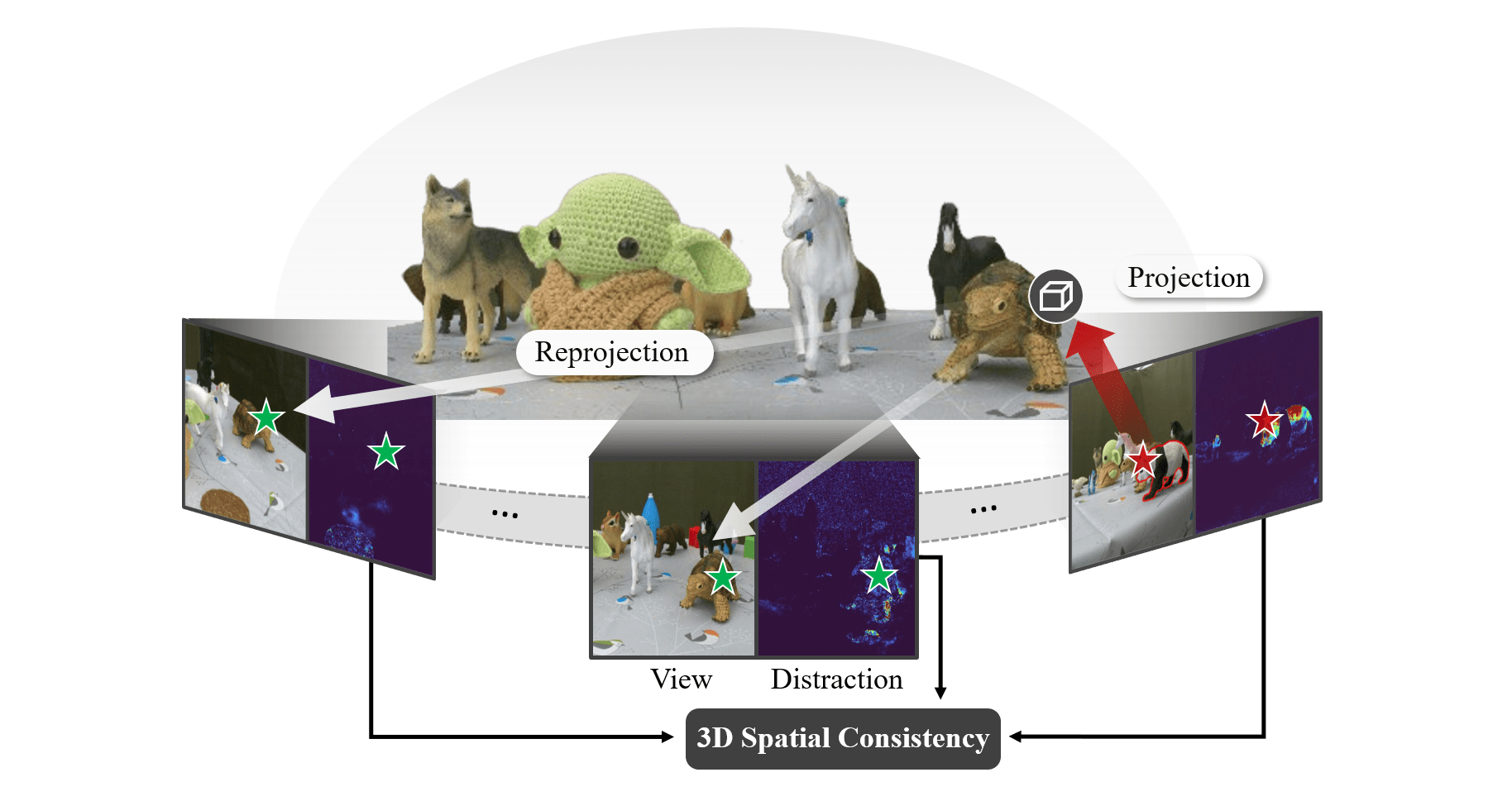}
    \end{center}
    \vspace{-0.1in}
    \caption{Illustration of PruNeRF, which includes measuring pixel-wise distraction scores and then assessing 3D spatial consistency using a depth-based reprojection technique. The red star and green stars denote a query pixel and projected pixels, respectively.}
    \label{fig:method-overview}
\end{figure*}

\section{Introduction}
Recent advancements in Neural Radiance Fields (NeRF)~\citep{nerf} have demonstrated remarkable performance in high-quality novel view synthesis using 2D supervision. However, achieving high quality in NeRF is significantly challenged by the need for well-curated multi-view images free from distractors which are unexpected objects present only in specific views (\emph{e.g.}, moving entities like pedestrians or birds). If certain pixels become corrupted due to distractors, it causes inconsistency in supervision, which inevitably results in reducing the model's performance. This issue contributes to the limited number of benchmark datasets in NeRF, especially compared with other domains such as 2D vision and point clouds. Constructing NeRF datasets typically requires capturing 2D images of a scene from multiple directions while maintaining consistent conditions, a constraint that complicates the process. However, despite its importance and timeliness, this challenge of handling distractors in datasets remains largely unexplored. 

A straightforward solution is a careful dataset curation to prevent the inclusion of distractors. However, in real-world scenarios, it is almost infeasible to preemptively inspect all training images for unknown distractors without prior knowledge about them. Therefore, simply allocating more resources for curation is not effective. Alternatively, recent studies have focused on developing a robust model designed for distinguishing transient objects from the target scene. NeRF-W~\citep{nerfw} uses a per-image latent code with an uncertainty-based network architecture to separate transient objects in uncontrolled web datasets, yet shows limited performance across datasets and has a hyperparameter selection issue. Several studies~\citep{nsff, d2nerf, dynamic} focus on continuously moving objects in monocular video but struggle to handle unordered images which are more common in real-world scenarios. 

More recently, RobustNeRF~\citep{robustnerf} demonstrated notable performance by applying a binary mask to pixel-wise loss during training. They generate the mask in each mini-batch through the intersection of three masks: the first selects pixels with the lowest 50\% of loss as inliers, the second diffuses this inlier mask to neighboring pixels, and the last refines the second one patch-wise. However, this approach has performance limitations due to its heuristic mask design and resilience on loss, which is not a precise metric. It also requires selecting hyperparameters for patch size and the threshold for patch-wise discrimination in the third mask. Furthermore, these specific training schemes are not compatible with other methods. For instance, integrating mask-based reweighting of this approach into uncertainty-based methods~\citep{nerfw, nerfvae} is challenging. 

In this paper, we propose PruNeRF, a segment-centric dataset pruning framework via 3D spatial consistency, that effectively identifies and prunes the distractors. Our method is motivated by human perception that humans recognize distractors by mapping objects from multiple images to virtual 3D space and assessing whether it is consistent or not. We begin by examining the performance of existing metrics for measuring pixel-wise distraction, such as loss and gradient, and introduce Influence Functions~\citep{inf_1974, inf_blackbox} that demonstrates more accurate precision. Unlike the previous metrics, which struggle to differentiate between hard-to-learn pixels and distractor pixels, we observed that Influence Functions exhibit superior performance in this regard. 

However, using these metrics alone is insufficient to solve the challenge considering inherent performance limitations and the need for manual threshold selection for accurate discrimination. Also, these metrics lack 3D awareness since they only measure general sample-wise distraction. Therefore, we propose 3D-aware distraction by assessing the 3D spatial consistency utilizing a depth-based reprojection technique. Similar to human perception, we project a query pixel onto a 3D surface point and then reproject it to corresponding pixels in other views. By estimating distributions of pixel-wise distraction scores, we identify pixels that significantly deviate from the derived distribution as distractors. 

Furthermore, to enhance the precision in identifying distractors at the segment level rather than the pixel level, we incorporate segmentation~\citep{kirillov2019panoptic}. Since distractions vary across pixels even within a single object, this approach significantly improves the identification. Unlike previous work~\citep{block}, our method does not necessitate a segmentation model trained for specific distractors. Instead, it suffices to partition the scene into smaller parts. We utilize SAM~\citep{sam} that demonstrates strong zero-shot capabilities for segmentation, enabling efficient use without additional training.

To this end, PruNeRF generates accurate masks to identify distractors from contaminated datasets. Considering the difficulties of constructing NeRF datasets, our method alleviates this burden by pruning unintended objects. Additionally, our method is compatible with various network architectures and can be integrated with a wide range of methods in a plug-and-play manner.

Our contributions are summarized as follows:
\begin{itemize}  
    \item We propose a novel 3D-aware distraction identification method that assesses 3D spatial consistency based on pixel-wise distraction.
     
    \item We introduce Influence Functions for measuring pixel-wise distraction in  NeRF, demonstrating superior performance compared with traditional metrics.
    
    \item We are the first to propose a dataset pruning approach that alleviates the challenges of dataset construction in NeRF. 
\end{itemize}

\section{Preliminaries}
\subsection{Neural Radiance Fields}
Neural Radiance Field represents a 3D scene as a continuous function $f_\theta$, implemented as a multi-layer perceptron (MLP) network. This function estimates RGB color $\mathbf{c}$ and volume density $\sigma$ from 3D coordinates $\mathbf{x}=(x, y, z)$ and 2D viewing direction $\mathbf{d}=(\theta, \phi)$ as $\{\mathbf{c}, \sigma\} = f_{\theta}(\mathbf{x}, \mathbf{d}).$ To ensure 3D-consistent geometry, the volume density $\sigma$ depends solely on the coordinates $\mathbf{x}$, making invariant to changes in viewing direction $\mathbf{d}$. Conversely, the RGB color $\mathbf{c}$ incorporates viewing direction $\mathbf{d}$ to account for non-Lambertian surfaces.

Given a camera ray $\mathbf{r}(t) = \mathbf{o} + t\mathbf{d}$, the corresponding pixel color $\hat{C}(\mathbf{r})$ is approximated by integrating the radiance within the interval from the camera's near plane $t_n$ to the far plane $t_f$:
\begin{equation*}
\begin{aligned}
\hat{C}(\mathbf{r}) & = \int_{t_n}^{t_f} T(t) \sigma(\mathbf{r}(t))\mathbf{c}\big(\mathbf{r}(t), \mathbf{d}\big)dt, \\
T(t) & =\exp\Big(-\int_{t_n}^{t} \sigma(\mathbf{r}(s)\Big)ds.
\end{aligned}
\end{equation*}
where $\mathbf{o}$ and $T(t)$ denote the camera center and the accumulated transmittance, respectively.

NeRF is optimized using a photometric loss with the ground truth pixel colors $C(\mathbf{r})$:
\begin{align}
\mathcal{L}(\mathcal{R}) = \sum_{\mathbf{r} \in \mathcal{R}} \big\|C(\mathbf{r}) - \hat{C}(\mathbf{r})\big\|_2^2.
\label{eq:nerfloss}
\end{align}
where $\mathcal{R}$ is the set of rays and $||\cdot\||_2$ denotes $L_2$-norm.. 

Given that the objective function of NeRFs does not account for distractors, the inclusion of distractors can easily degrade performance by compromising the ground truth 3D consistency.

\subsection{Anomaly Measurement Metrics}
\label{subsec:if}
There have been several attempts to understand training samples using information from the output space or parameter space of trained models in deep learning~\citep{odin, mahalanobis, gradnorm, gradcam}. 

In empirical risk minimization (ERM), the goal is to minimize the finite-sum objective for a trainset $D= \{z_i : (x_{i}, y_{i})\}_{i=1}^{N}$ as ${\theta}^*= \text{argmin}_{\theta} \mathcal{L}(D, \theta)$, where $\theta\in\mathbb{R}^P$ represents the network parameter and $\mathcal{L}$ is the sum of sample-wise loss over the dataset:
$\mathcal{L}(D,\theta) = \frac{1}{N}\sum_{i=1}^N\ell(z_i, \theta)$, where $\ell$ denotes the sample-wise loss function. 

\vspace{-0.05in}
\paragraph{Loss} The sample-wise loss using information from the trained model's output space can be described as $\ell(z_i, \hat{\theta})$, where $\hat{\theta}$ is the parameter of the trained network. Since minority samples tend to exhibit high values due to marginalization during training, this metric is widely utilized for its simplicity and computational efficiency. However, the inherent tendency of neural networks to memorize training samples can diminish its efficacy as training progresses~\citep{memorization, memorization2}.

\vspace{-0.05in}
\paragraph{Gradient} Sample-wise gradient-norm with respect to the parameters of the trained model is widely used for tasks such as out-of-distribution detection~\citep{gradnorm, gradnorm2}. Gradient-norm is represented as $||\nabla_{\theta}\ell(z_i, \hat{\theta})||_p$, where $||\cdot\||_p$ denotes $L_p$-norm.

\begin{figure*}[t]
    \begin{center}
    \begin{subfigure}[t]{.24\linewidth}
    \centering\includegraphics[width=1.0\linewidth, height=3.5cm]{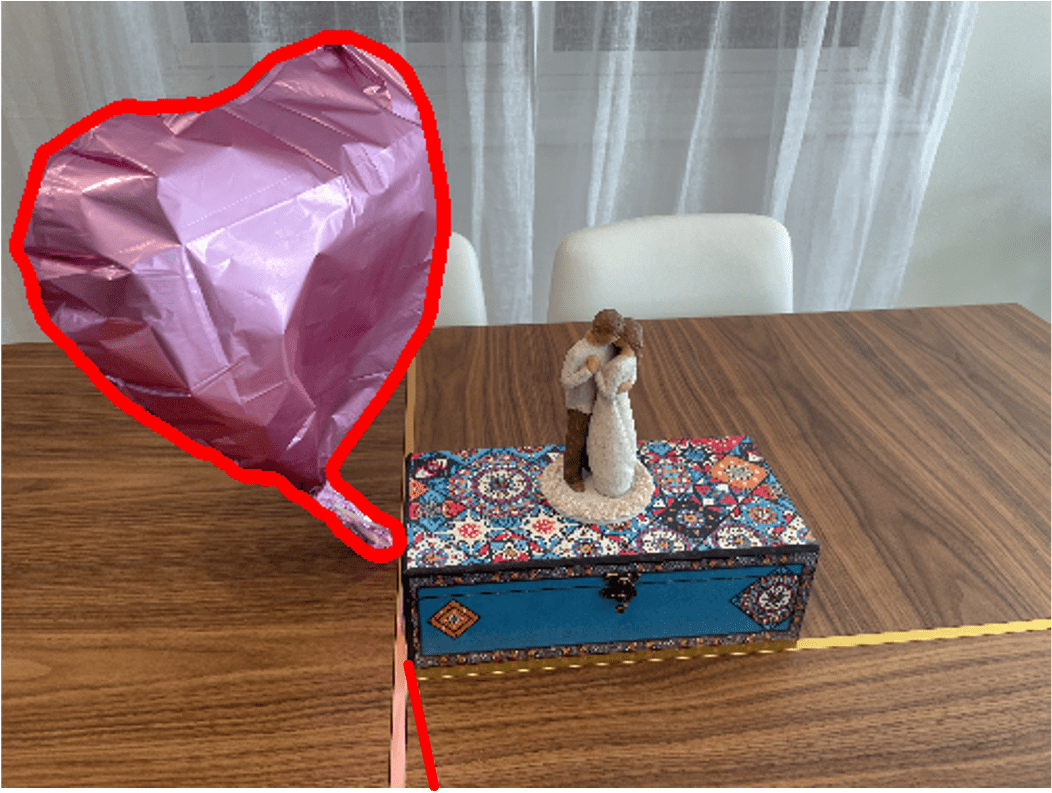}
    \end{subfigure}
    \begin{subfigure}[t]{.24\linewidth}
    \includegraphics[width=1.0\linewidth, height=3.5cm]
    {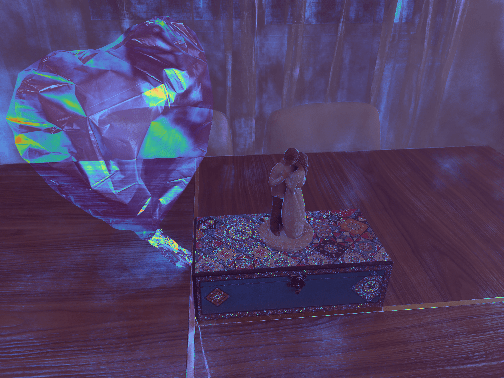}
    \end{subfigure}
    \begin{subfigure}[t]{.24\linewidth}
    \includegraphics[width=1.0\linewidth, height=3.5cm]    {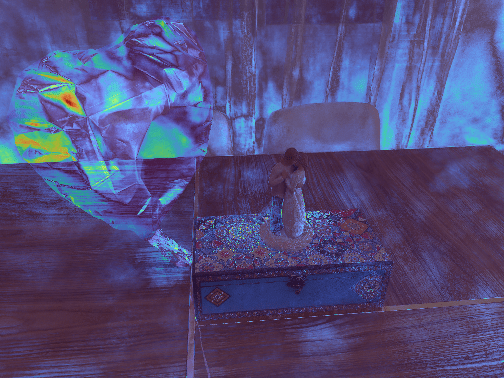}
    \end{subfigure}
    \begin{subfigure}[t]{.24\linewidth}
    \includegraphics[width=1.0\linewidth, height=3.5cm]{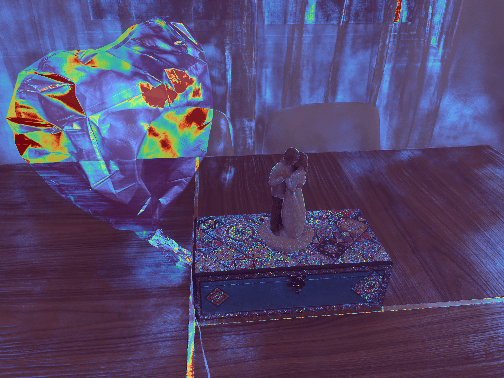}
    \end{subfigure}
    \end{center}

    \begin{center}
    \begin{subfigure}[t]{.24\linewidth}
    \includegraphics[width=1.\linewidth, height=3.5cm]{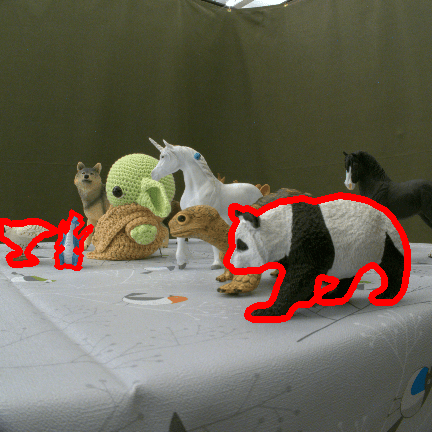}
    \caption{Training Image}
    \end{subfigure}
    \begin{subfigure}[t]{.24\linewidth}
    \includegraphics[width=1.\linewidth, height=3.5cm]{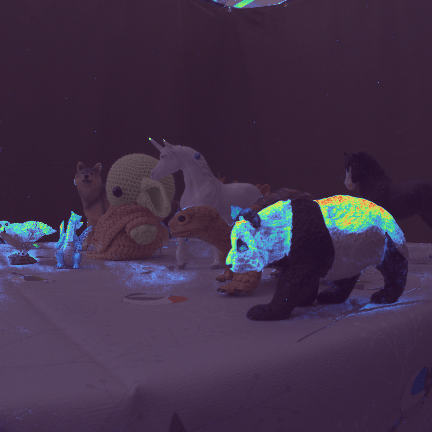}
    \caption{Loss}
    \label{fig:loss}
    \end{subfigure}
     \begin{subfigure}[t]{.24\linewidth}
    \includegraphics[width=1.\linewidth, height=3.5cm]{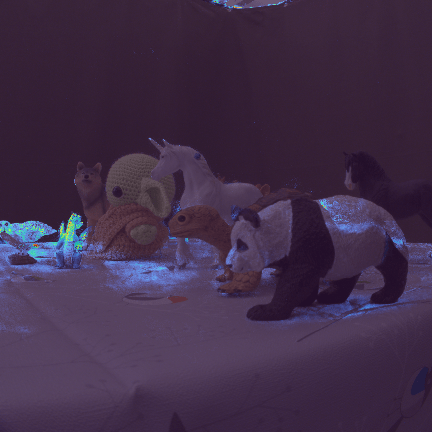}
    \caption{GradNorm}
    \label{fig:gradnorm}
    \end{subfigure}
    \begin{subfigure}[t]{.24\linewidth}
    \includegraphics[width=1.\linewidth, height=3.5cm]{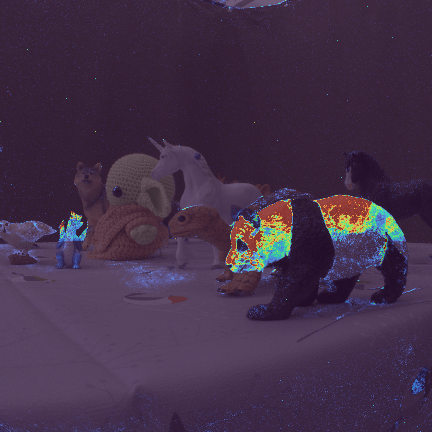}
    \caption{Influence Function}
    \label{fig:if}
    \end{subfigure}
    \end{center}
    \vspace{-0.1in}
    \caption{Comparison of pixel-wise measurements in two natural scenes: Statue and Baby Yoda. In (a), the boundaries of distractors are highlighted in red. In (b) and (c), the heatmaps of loss and gradient-norm are relatively noisy and struggle to discern distractors from hard-to-learn regions. However, in (d), Influence Function reveals a more precise detection of distractors compared with others.}
    \label{fig:metrics}
\end{figure*}

\vspace{-0.05in}
\paragraph{Influence Functions}
Another approach for estimating the influence of a data point is Leave-One-Out (LOO) retraining, which involves excluding a data point from the trainset, retraining the model, and assessing the impact of this sample by comparing the model's predictions before and after the exclusion.

The estimated impact of a data point $z_i$ on another sample $z_j$ using LOO retraining is defined as 
\begin{equation*}
    \mathcal{I}_{LOO}(z_i,z_j) \approx \ell(z_j,\theta^*_{-z_i}) - \ell(z_j,\theta^*).
\end{equation*}
where ${\theta^{*}_{-z_i}}$ is retrained parameter without sample $z_i$. However, conducting LOO retraining for each data point is computationally infeasible.

To alleviate prohibitive computational costs, Influence Functions~\citep{inf_blackbox} are proposed as an approximation of LOO retraining by approximating the parameter difference as a Newton ascent term:
\begin{align}
    \mathcal{I}(z_i,z_j)
    & :=
    \nabla_{\theta}\ell(z_j,\theta^*)^\top \mathbf{H}^{-1} \nabla_{\theta}\ell(z_i,\theta^*).
    \label{eq:inf}
\end{align}
where $\mathbf{H}= \nabla_{\theta}^{2} \mathcal{L}(D, \theta^*) \in \mathbb{R}^{P \times P}$ is the Hessian of the loss function with respect to the parameter $\theta^*$ and is positive semi-definite since $\theta^*$ is a local minimum.

Typically, if a validation set is available, influence can be calculated as $\mathcal{I}(z_i, \mathcal{V})$, where $\mathcal{V}$ represents the validation set. However, in practical scenarios, a clean validation set is often unavailable. In such cases, Self-Influence $\mathcal{I}(z_i, z_i)$ is used to reflect the model's predictive capability on $z_i$ when trained without it. A higher Self-Influence score indicates more difficulty in prediction, indicating that $z_i$ significantly deviates from the majority of the trainset.

\section{Method}
In this section, we propose PruNeRF, a segment-centric dataset pruning framework for NeRF that leverages 3D spatial consistency. Our method identifies distractors in the trainset by first measuring pixel-wise distraction and then assessing 3D spatial consistency. Through a pixel-to-segment refinement based on a segmentation model, we accurately prune distractors at the object level. 

In \cref{subsec:inf}, we examine the performance of existing metrics for measuring pixel-wise distraction and introduce Influence Functions demonstrating its effectiveness in providing more accurate measurements. In \cref{subsec:consistency}, we quantify 3D-aware distraction by assessing 3D spatial consistency based on the measured pixel-wise distraction. In addition, we incorporate segmentation for pixel-to-segment refinement, enabling more accurate identification of distractors in \cref{subsec:seg}. Our method is outlined in \cref{fig:method-overview}.

\begin{figure*}[t]
    \begin{center}
    \includegraphics[width=0.19 \textwidth, height=3cm]{{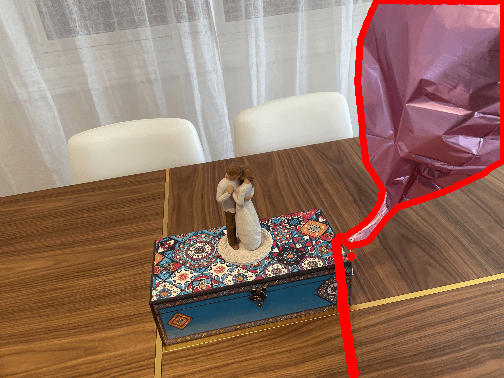}}
    \includegraphics[width=0.19 \textwidth, height=3cm]{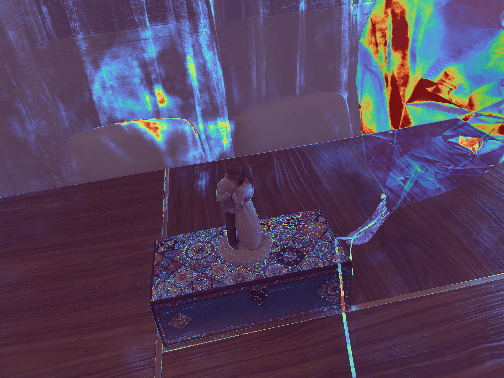}
    \includegraphics[width=0.19 \textwidth, height=3cm]{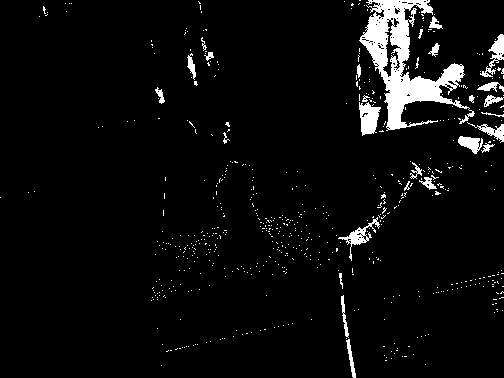}
    \includegraphics[width=0.19\textwidth, height=3cm]{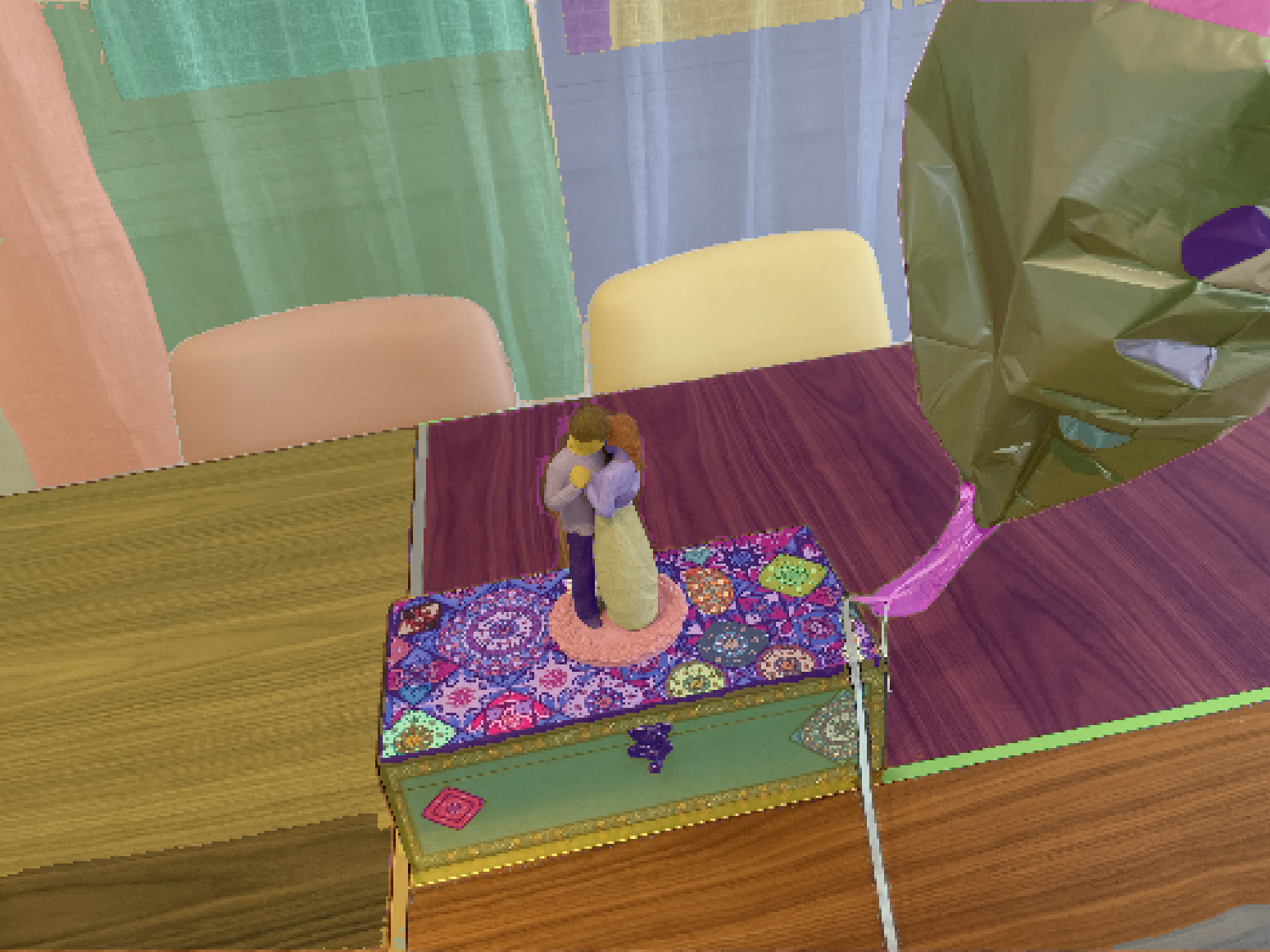}
    \includegraphics[width=0.19 \textwidth, height=3cm]{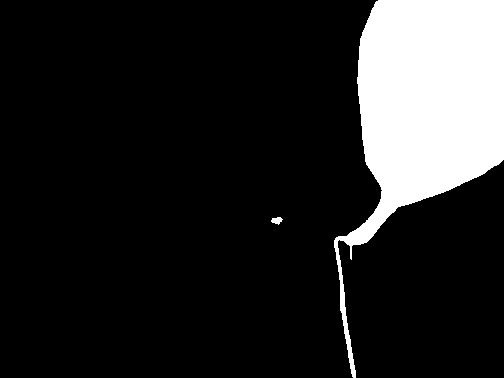}
    \end{center}
    
    \begin{center}
    \includegraphics[width=0.19\textwidth, height=3cm]{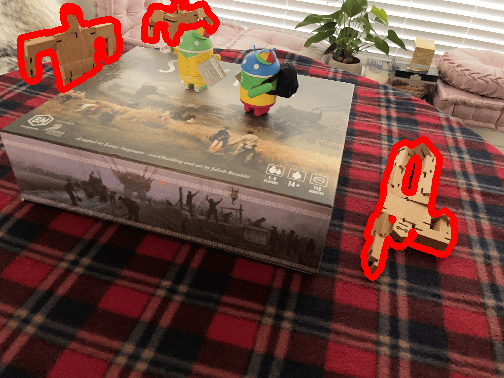}
    \includegraphics[width=0.19\textwidth, height=3cm]{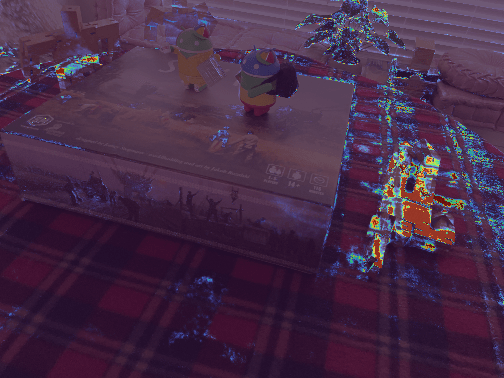}
    \includegraphics[width=0.19\textwidth, height=3cm]{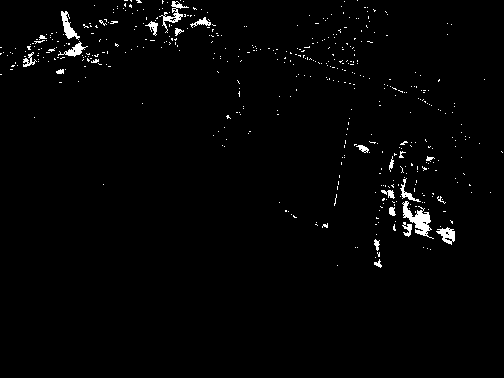}
    \includegraphics[width=0.19\textwidth, height=3cm]{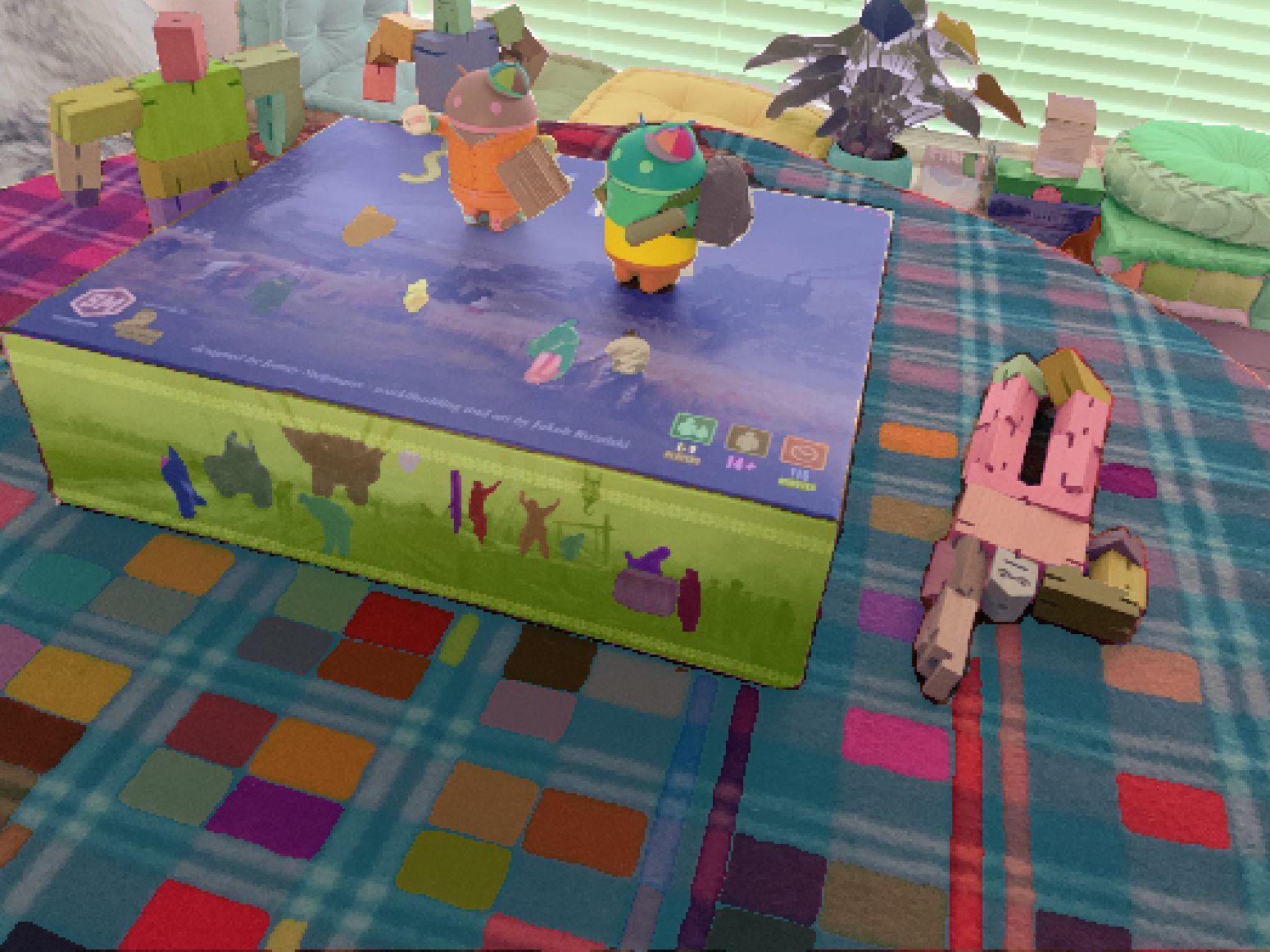}
    \includegraphics[width=0.19\textwidth, height=3cm]{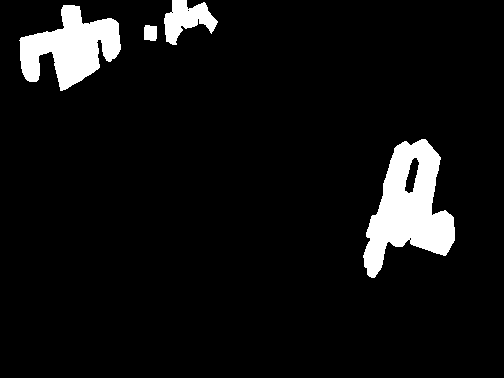}
    \end{center}
    
    \begin{center}
    \begin{subfigure}[t]{.19\textwidth}
    \includegraphics[width=1.\textwidth, height=3cm]{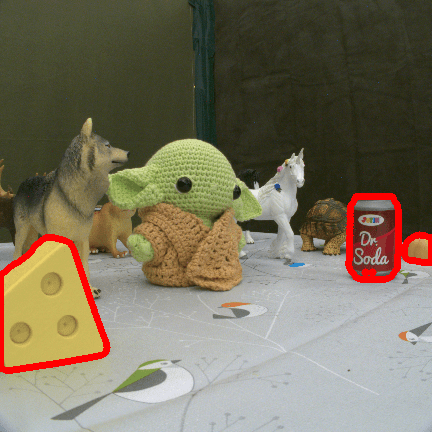}
    \caption{Training Image}
    \end{subfigure}
    \begin{subfigure}[t]{.19\textwidth}
    \includegraphics[width=1.\textwidth, height=3cm]{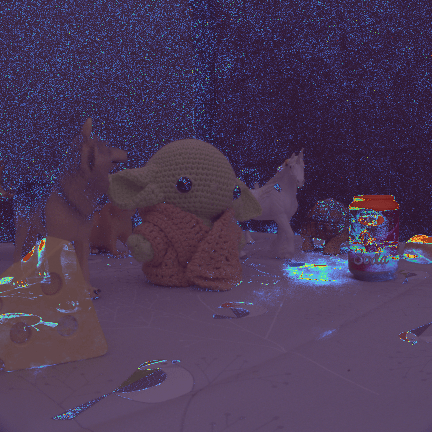}
    \caption{Influence Function}
    \end{subfigure}
    \begin{subfigure}[t]{.19\textwidth}
    \includegraphics[width=1.\textwidth, height=3cm]{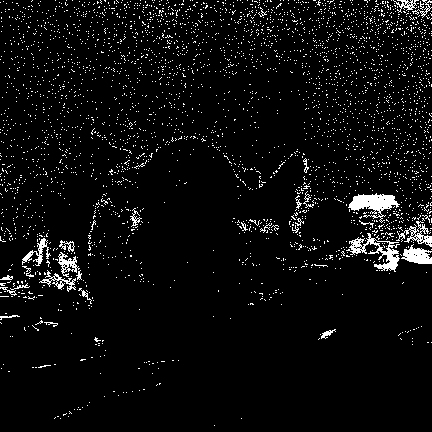}
    \caption{3D Consistency}
    \end{subfigure}
    \begin{subfigure}[t]{.19\textwidth}
    \includegraphics[width=1.\textwidth, height=3cm]{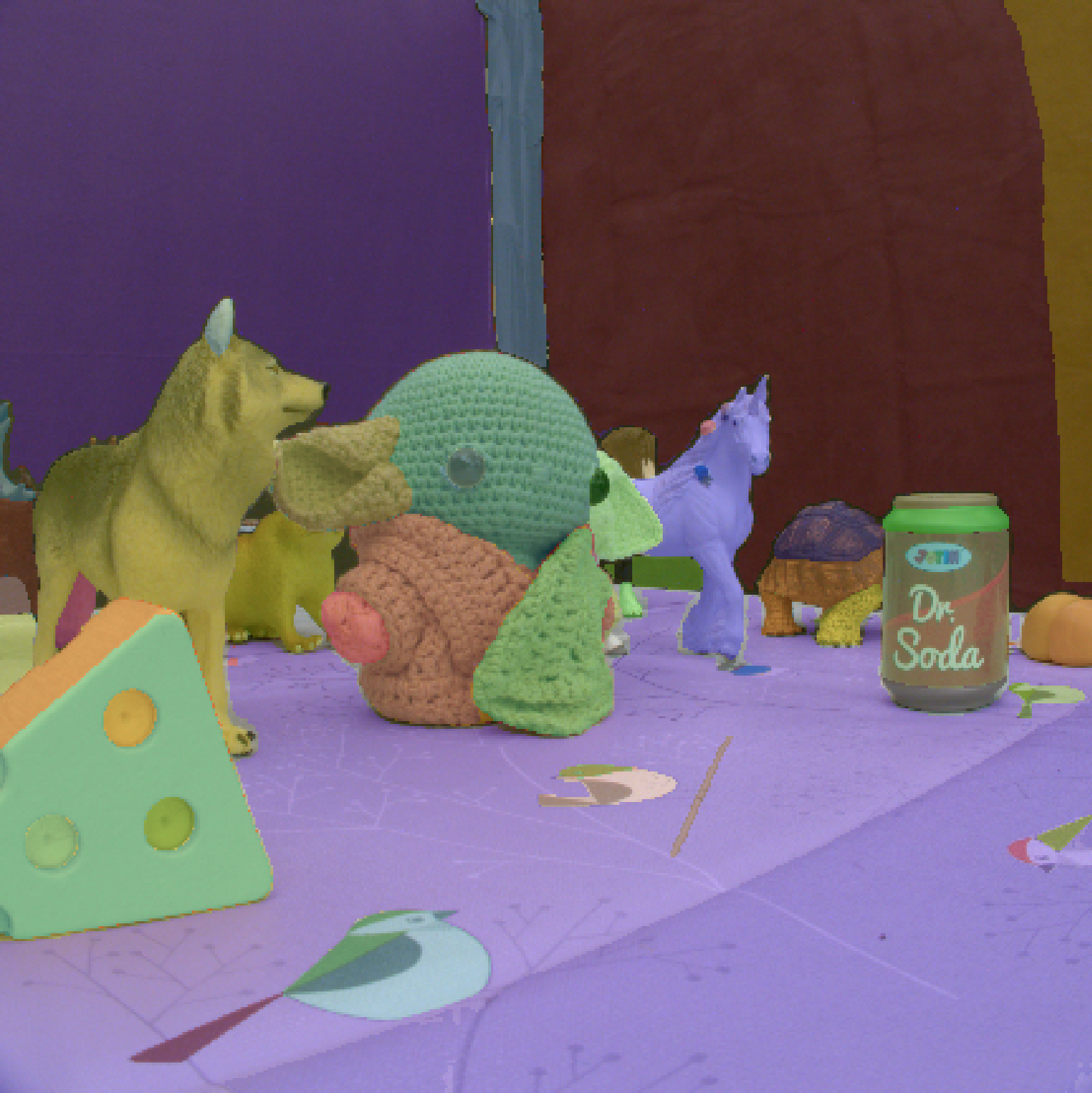}
    \caption{Segmentation}
    \label{fig:step_seg}
    \end{subfigure}
    \begin{subfigure}[t]{.19\textwidth}
    \includegraphics[width=1.\textwidth, height=3cm]{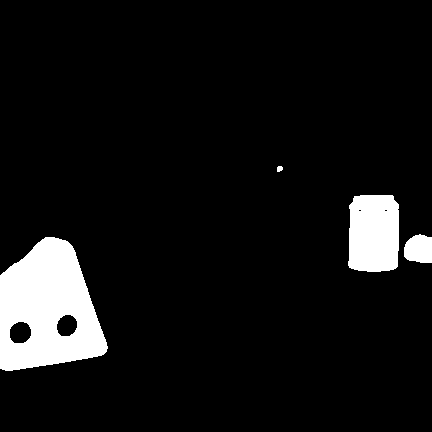}
    \caption{Pixel-to-Segment}
    \end{subfigure}
    \end{center}
    \vspace{-0.15in}
    \caption{Visualization of each step in PruNeRF. (a) Training image with distractors highlighted in red. (b) Influence heatmaps identifying distractor pixels (\cref{subsec:if}). (c) 3D spatial consistency further refines distractor identification (\cref{subsec:consistency}). (d) and (e) show the segmentation results and our final mask of distractors through pixel-to-segment refinement, respectively (\cref{subsec:seg}).}
    \label{fig:step}
\end{figure*}

\subsection{Measuring Pixel-wise Distraction}
\label{subsec:inf}
In the training phase of NeRF, we typically focus on minimizing the finite-sum objective function, similar to other deep learning tasks, as defined in \cref{eq:nerfloss}. This process enables the model to learn the 3D geometry, implicitly guided by the ground truth pixels from multi-view training data. However, the presence of distractors in the trainset continuously provides malicious signals that prevent the model from learning accurate 3D geometry during training. As a result, some distractor pixels achieve lower loss values through the minimization of the objective function, while others, contradicted by the predominant signals from the majority of pixels representing the target scene, exhibit high loss values. 

In this context, it is reasonable to approach identifying distractor pixels using pixel-wise loss derived from a trained model. Additionally, given the infeasibility of preemptively inspecting all training images for unknown distractors, it is practical to leverage a trained model that has been contaminated by distractors in the trainset. To verify the effectiveness of metrics based on a trained model's output and parameter space, such as loss and gradient, we quantify pixel-wise distraction using these metrics on the benchmark datasets, Statue and BabyYoda, where a balloon and various arbitrary objects appear in random regions as distractors in each view, respectively. Detailed descriptions of the datasets are provided in \cref{appendix:dataset}.

In \cref{fig:loss} and \cref{fig:gradnorm}, when using loss or gradient-norm, some distractor pixels show high values while others do not within distractors. Moreover, these metrics exhibit high values in hard-to-learn regions, such as the semi-transparent curtain and tablecloth patterns, in each dataset. Considering that neural networks tend to memorize outlier samples, such as mislabeled ones, due to the repeated minimization of the objective function~\citep{memorization, memorization2}, these metrics alone are insufficient for identifying distracting pixels.

To achieve better performance, we introduce Influence Functions~\citep{inf_blackbox} that measure the influence of individual train samples on each other. In scenarios where a model is trained with a normal dataset, even if some pixels are excluded from the trainset, the model can still predict them accurately by using the knowledge from other pixels.
Conversely, when training a model with distractors, excluding distractor pixels results in a noticeable decrease in prediction performance on these pixels, as normal pixels cannot provide information about distractors due to inconsistency. This implies that including distractor pixels in the trainset is crucial for the model to accurately infer them. Moreover, since NeRF learns 3D space through the strong interactions between data samples, this behavior is more pronounced than 2D vision.

Motivated by this, we employ Influence Functions to efficiently approximate Leave-One-Out retraining for identifying pixels that deviate from the overall dataset consistency. If a clean validation set is available, we can calculate the influence of a pixel ${z}_i$ on a validation point $\mathcal{V}$ as $\mathcal{I}({z}_i, \mathcal{V})$ with \cref{eq:inf}. However, it is generally impractical to preemptively curate such a validation set in real-world scenarios. Therefore, we calculate Self-Influence score $\mathcal{I}({z}_i, {z}_i)$. It quantifies the influence on the model's prediction for a pixel ${z}_i$ (for a ray $\mathbf{r}_i \in \mathcal{R}$) when a pixel itself is excluded from the trainset. Note that distractor pixels tend to show high scores since clean pixels from most of the trainset cannot provide the information needed to predict these out-of-distribution distractor pixels.

\renewcommand{\arraystretch}{1.3}
\begin{table}[t]
    \centering
    \caption{Performance comparison of distraction measuring metrics using top-k\% pruning, with the average PSNR on test sets for $k$ of 5\% and 10\%.}
    \vspace{-0.05in}
    \setlength{\tabcolsep}{5pt}
    \resizebox{1.0\linewidth}{!}{
        \begin{tabular}{lcccccc}
        \toprule
        \multirow{2}{*}{Metric} & \multicolumn{2}{c}{Statue} & \multicolumn{2}{c}{Android} & \multicolumn{2}{c}{BabyYoda} \\
        \cline{2-3} \cline{4-5} \cline{6-7}     
        & 5\% & 10\% & 5\% & 10\% & 5\% & 10\%\\
        \midrule 
        Loss & 19.87 & 19.76 & 22.20 & 22.18 & 30.37 & 30.46\\
        GradNorm & 19.96 & 19.88 & 21.82 & 21.78 & 30.68 & 30.68\\
        IF & \textbf{19.98} & \textbf{20.29} & \textbf{22.32} & \textbf{22.42} & \textbf{30.83} & \textbf{30.98} \\
        \bottomrule
        \label{table:topk}
        \end{tabular}
    }
\vspace{-0.3in}
\end{table}

To quantitatively compare the efficacy of these metrics in mitigating distractors, we measure the PSNR on the test set using a model trained on datasets pruned by the top-$k$\% of pixels based on loss, gradient-norm, and Influence Function. Since the optimal $k$ is unknown, we conduct experiments using {5\%, 10\%}. In addition, we use the parameters of the last layer of the model for calculating the Hessian inverse $H^{-1}$ and the loss gradient to reduce the computational cost associated with network parameters following \citet{inf_blackbox, tracin}. As shown in \cref{table:topk}, Influence Function achieves higher performance than others in natural scenes. Additionally, \cref{fig:if} demonstrates that Influence Function excels in highlighting distracting regions compared with loss and gradient-norm. These results support the efficacy of using Influence Function as the base measure for identifying distractors in our method. Notably, our method has the advantage of being able to incorporate any metrics that measure sample-wise distraction, including loss and gradient-norm, in a plug-and-play manner.

It is also important to highlight that the top-$k$\% pruning approach poses a challenge. This challenge arises from the significant cost required to search the optimal hyperparameter $k$ for each dataset, particularly given the unknown types and numbers of distractors. This underscores the need for an efficient component that is generally adaptable to various datasets.

\subsection{Distractor Detection with 3D Spatial Consistency}
\label{subsec:consistency}
Due to the lack of consideration of 3D space in the aforementioned metrics including Influence Function, relying solely on these metrics is insufficient to address the problem effectively. Therefore, we propose 3D-aware distraction by assessing 3D spatial consistency motivated by human perception. Our method is inspired by human perception that humans identify distractors by simultaneously mapping objects from multiple images to virtual 3D space and verifying whether it is consistent or not. We employ a depth-based reprojection technique based on measured pixel-wise distraction, mirroring the human behavior to achieve 3D-aware distractor identification

Specifically, we estimate the depth of a query pixel using a trained NeRF to locate a 3D point on the surface. Then, we project this 3D point to other views to identify the corresponding pixels. To mitigate occlusion and depth error issues, we estimate depth points at both the source and projected viewpoints and use points whose Euclidean distance between them falls below a threshold $\theta$. We set $\theta$ to 0.1 for all datasets. This reprojection process enables us to estimate a distribution of Self-Influence scores for corresponding pixels that view the same 3D point.

Since distracting pixels yield high influence scores, we identify pixels that significantly deviate from the derived distribution as distractors. To measure this deviation, we first compute the mean and standard deviation of this distribution for each ray $\mathbf{r}_i$ as
\vspace{-0.1in}
\begin{align*}
    &\mu_{P_{\mathbf{r}_i}} = \frac{1}{|P_{\mathbf{r}_i}|}\sum_{{\mathbf{r}_j} \in P_{\mathbf{r}_i}} \mathcal{I}(\mathbf{r}_j, \mathbf{r}_j), \\
    &\sigma_{P_{\mathbf{r}_i}} = \sqrt{\frac{1}{|P_{\mathbf{r}_i}|}\sum_{{\mathbf{r}_j} \in P_{\mathbf{r}_i}} \left(\mathcal{I}(\mathbf{r}_j, \mathbf{r}_j) - \mu_{P_{\mathbf{r}_i}}\right)^2}
\end{align*}

Following the classical outlier detection approach, we define the indicator function $\mathbf{1}_{P_{\mathbf{r}_i}}:\mathcal{R}\rightarrow \{0,1\}$ as
\vspace{-0.1in}
\begin{equation*}
    \mathbf{1}_{P_{\mathbf{r}_i}}(\mathbf{r}_j) =
        \begin{cases}
          1 & \text{if  $\mathcal{I}(\mathbf{r}_j,\mathbf{r}_j) \notin
            (\mu-3\sigma, \mu+3\sigma)$}, \\
          0 & \text{otherwise}.
        \end{cases}
\vspace{-0.1in}
\end{equation*}
where $\mu$ and $\sigma$ denote $\mu_{P_{\mathbf{r}_i}}$ and $\sigma_{P_{\mathbf{r}_i}}$, respectively. Note that the value is 1 if $\mathbf{r}_j$ is a distractor pixel.

\begin{figure*}[t]
    \begin{center}
    \includegraphics[width=0.24 \textwidth, height=3cm]{{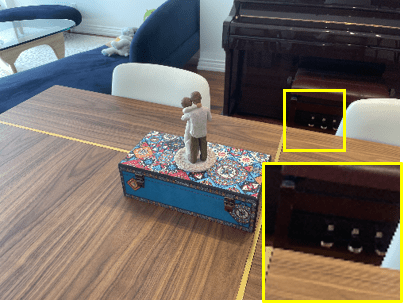}}
    \includegraphics[width=0.24 \textwidth, height=3cm]{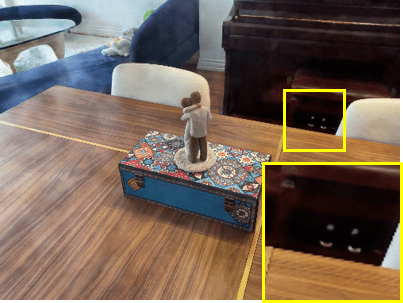}
    \includegraphics[width=0.24 \textwidth, height=3cm]{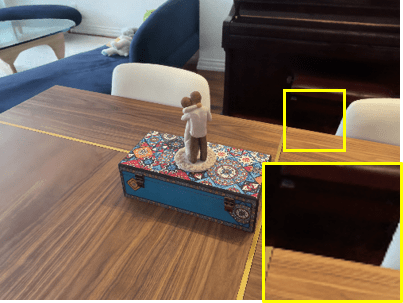}
    \includegraphics[width=0.24 \textwidth, height=3cm]{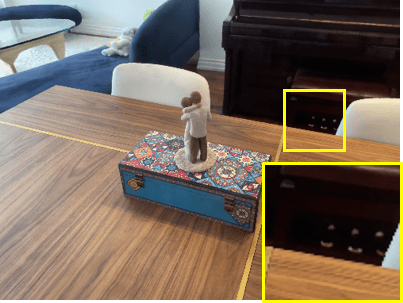}
    \end{center}
    
    \begin{center}
    \includegraphics[width=0.24 \textwidth, height=3cm]{{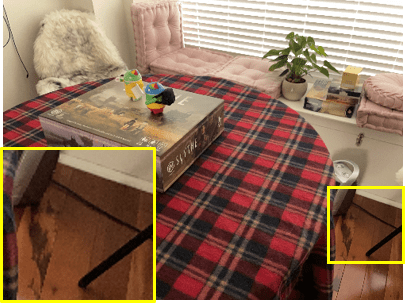}}
    \includegraphics[width=0.24 \textwidth, height=3cm]{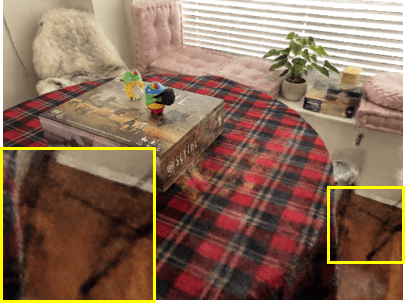}
    \includegraphics[width=0.24 \textwidth, height=3cm]{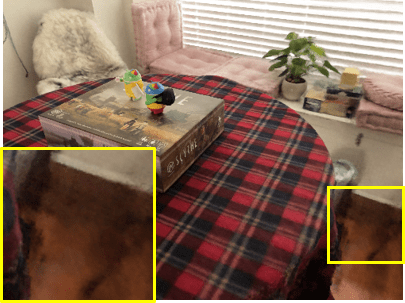}
    \includegraphics[width=0.24 \textwidth, height=3cm]{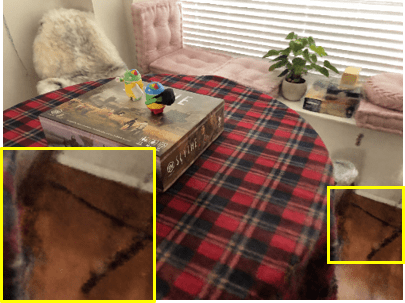}
    \end{center}

    \begin{center}
    \begin{subfigure}[t]{0.24\textwidth}
    \includegraphics[width=1.\textwidth, height=3cm]{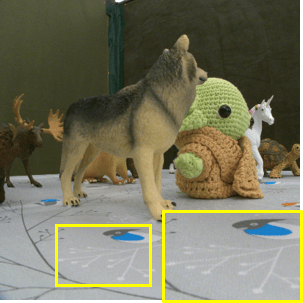}
    \caption{Ground Truth}
    \end{subfigure}
    \begin{subfigure}[t]{0.24\textwidth}
    \includegraphics[width=1.\textwidth, height=3cm]{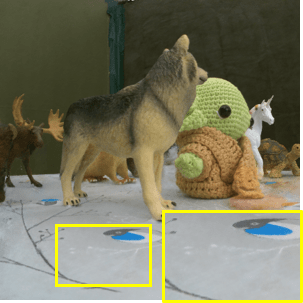}
    \caption{MipNeRF360}
    \end{subfigure}
    \begin{subfigure}[t]{.24\textwidth}
    \includegraphics[width=1.\textwidth, height=3cm]{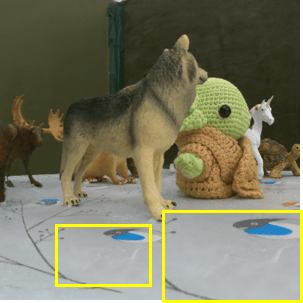}
    \caption{RobustNeRF}
    \end{subfigure}
    \begin{subfigure}[t]{.24\textwidth}
    \includegraphics[width=1.\textwidth, height=3cm]{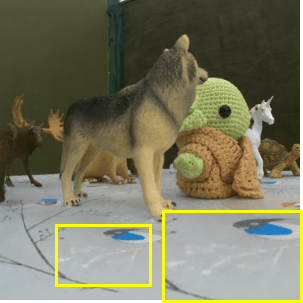}
    \caption{PruNeRF}
    \end{subfigure}
    \end{center}
    \vspace{-0.15in}
    \caption{Qualitative evaluation on Statue, Android, and BabyYoda. PruNeRF exhibits superior performance, especially in hard-to-learn regions highlighted in the yellow box. Additional results are provided in \cref{appen:qual_eval}.}
    \label{fig:qual_eval1}
\end{figure*}

\begin{figure*}[t]
    \begin{center}
    \begin{subfigure}[t]{0.19\textwidth}
    \includegraphics[width=1.\textwidth, height=3cm]{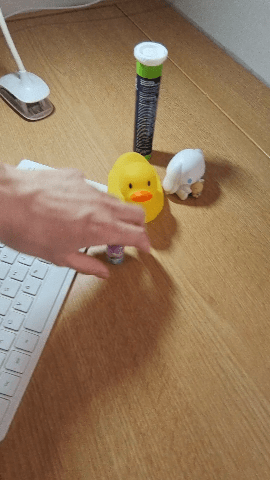}
    \caption{Training Image}
    \end{subfigure}
    \begin{subfigure}[t]{0.19\textwidth}
    \includegraphics[width=1.\textwidth, height=3cm]{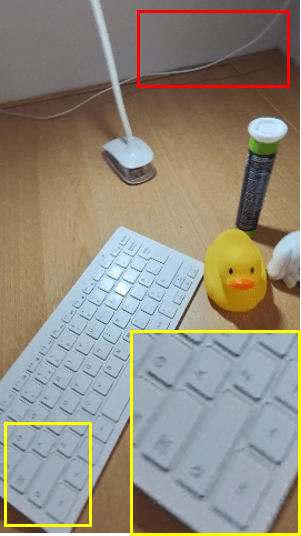}
    \caption{Ground Truth}
    \end{subfigure}
    \begin{subfigure}[t]{.19\textwidth}
    \includegraphics[width=1.\textwidth, height=3cm]{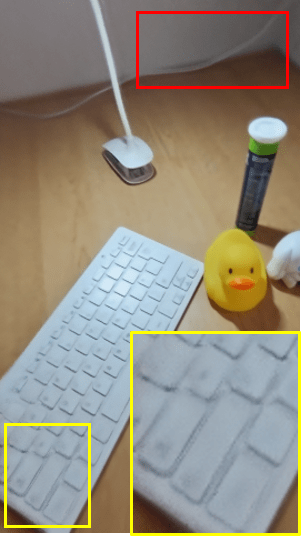}
    \caption{D2NeRF}
    \end{subfigure}
    \begin{subfigure}[t]{.19\textwidth}
    \includegraphics[width=1.\textwidth, height=3cm]{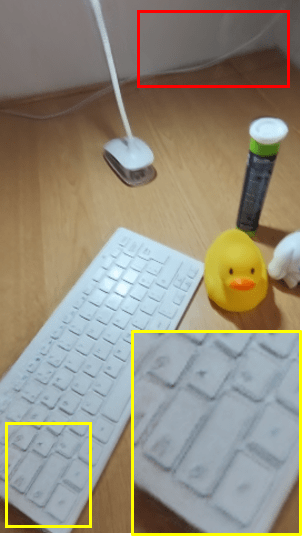}
    \caption{RobustNeRF}
    \end{subfigure}
    \begin{subfigure}[t]{.19\textwidth}
    \includegraphics[width=1.\textwidth, height=3cm]{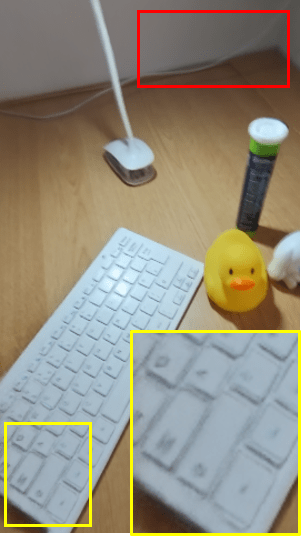}
    \caption{PruNeRF}
    \end{subfigure}
    \end{center}
    \caption{Qualitative evaluation on Pick. This dataset includes a moving hand as a distractor that changes its location with temporal continuity. PruNeRF demonstrates enhanced performance, particularly in hard-to-learn regions as emphasized in the box, compared with the state-of-the-art methods.}
    \label{fig:qual_eval_d2}
\end{figure*}

\subsection{Pixel-to-Segment Refinement}
\label{subsec:seg}

To further enhance the detection accuracy of distractors, we propose a pixel-to-segment refinement method. As observed in \cref{fig:metrics}, distractions vary across pixels even within a single object. This variability makes pixel-level pruning prone to noise and inaccuracies. To address this issue, our method integrates segmentation to improve precision by identifying distractors at the segment level instead of the pixel level. Specifically, we employ SAM~\citep{sam} for refinement owing to its remarkable zero-shot partitioning performance, eliminating the need for additional training. Notably, any segmentation model capable of partitioning the scene into small segments can be seamlessly incorporated into our method.

Leveraging the segmentation model that maps each pixel (ray $\mathbf{r}_i$) to a semantic class $c \in \mathcal{C}$ and an instance id $z \in \mathbb{N}$, we compute the ratio of distracting pixels segment-wise to refine the mask $\mathbf{1}_{S}(S_{c,z})$ as follows:
\begin{equation*}
    \mathbf{1}_{S}(S_{c,z}) =
        \begin{cases}
          1 & \text{if $\left\{\frac{1}{|S_{c,z}|}\sum_{\mathbf{r}_j \in S_{c,z}} \mathbf{1}_{P_{\mathbf{r}_i}}(\mathbf{r}_j)\right\} > \epsilon$}, \\
          0 & \text{otherwise}
        \end{cases}
\end{equation*}
where $S_{c,z}$ is a set of pixels in the segment with a semantic class $c$ and an instance id $z$, and $\epsilon$ is a threshold. Threshold $\epsilon$ is used to filter out segments containing a small number of noisy pixels. Although dataset-specific tuning could offer further improvements, we set $\epsilon$ to 0.1 consistently across all datasets, as our target problem involves unknown types and quantities of distractors. In \cref{fig:step}, we present the results from each step of our method, showing that performance improves with each additional step. 

In our method, the prerequisite for a segmentation model is not to segment objects semantically (\emph{e.g.}, a pedestrian or a table) but rather to partition the scene into sufficiently small parts (\emph{e.g.}, a leg of a table). While a semantic segmentation model trained with specific distractor classes would be ideal~\citep{block}, it is prohibitively costly, and acquiring prior knowledge about distractors and sufficient multi-view data for training is challenging. 

\renewcommand{\arraystretch}{1.3}
\begin{table*}[t]
    \centering
    \caption{Quantitative evaluation on the synthetic scenes.}
    \setlength{\tabcolsep}{1pt} 
    \resizebox{1.00\linewidth}{!}{
    \begin{tabular}{@{\extracolsep{4pt}}lcccccccccc@{}}
    \toprule
    \multirow{2}{*}{\textbf{Method}} & \multicolumn{2}{c}{Car} & \multicolumn{2}{c}{Cars} & \multicolumn{2}{c}{Bag} & \multicolumn{2}{c}{Chairs} & \multicolumn{2}{c}{Pillow} \\
    \cline{2-3} \cline{4-5} \cline{6-7} \cline{8-9} \cline{10-11}    
    & MS-SSIM $\uparrow$ & PSNR $\uparrow$ & MS-SSIM $\uparrow$ & PSNR $\uparrow$ & MS-SSIM $\uparrow$ & PSNR $\uparrow$ & MS-SSIM $\uparrow$ & PSNR $\uparrow$ & MS-SSIM $\uparrow$ & PSNR $\uparrow$ \\
    \midrule 
    NeRF-W & .814 & 24.23 & .873 & 24.51 & .791 & 20.65 & .681 & 23.77 & .935 & 28.24\\
    NSFF & .806 & 24.90 & .376 & 10.29 & .892 & 25.62 & .284 & 12.82 & .343 & 4.55\\
    NeuralDiff & .952 & 31.89 & .921 & 25.93 & .910 & 29.02 & .722 & 24.42 & .652 & 20.09\\
    D$^2$NeRF & .975 & 34.27 & .953 & 26.27 & .979 & 34.14 & .707 & 24.63 & .979 & 36.58\\
    RobustNeRF & .992 & 38.17 & .982 & 27.32 & .994 & 39.34 & .987 & 37.82 & .995 & 39.85\\
    PruNeRF & \textbf{.994} & \textbf{39.19} & \textbf{.982} & \textbf{27.36} & \textbf{.996} & \textbf{41.23} & \textbf{.990} & \textbf{38.91} & \textbf{.996} & \textbf{41.21} \\
    \bottomrule
    \label{table:synthetic}
    \end{tabular}
    }
\vspace{-0.2in}
\end{table*}

To this end, PruNeRF generates masks that identify distractors from contaminated datasets. Given the challenges of dataset construction in NeRF, such as capturing scenes from diverse angles while controlling external factors, our dataset pruning approach can alleviate this burden by pruning accidentally included objects. Furthermore, our method is compatible with various network architectures and can be integrated with other methods in a plug-and-play manner.

\section{Experiments}
In this section, we evaluate our method across various benchmark datasets to assess its robustness against the presence of distractors. We describe our experimental setup in \cref{subsec:exp_setup}. Then, we present the evaluation results in \cref{subsec:results} and the ablation study in \cref{subsec:ablation}. 

\subsection{Experimental Settings}
\label{subsec:exp_setup}
\paragraph{Datasets}
We conduct experiments on benchmark datasets, including both synthetic and natural datasets, to evaluate the effectiveness of robustness against distractors. For the synthetic datasets, we utilize Kubric datasets~\citep{d2nerf}, which consist of video frames with moving objects that continuously move within specific scenes. These datasets comprise five types: car, cars, bag, chairs, and pillow. Each dataset includes 200 temporally ordered views for the train set and 100 novel views for the test set. For the natural datasets, we use Pick~\cite{d2nerf} and {Statue, Android, BabyYoda}~\citep{robustnerf}. Pick includes a moving hand as a distractor that changes its location with temporal continuity. Statue contains a single balloon distractor that changes location slightly with each view. Android includes three small wooden robots positioned in entirely new locations in each view. Yoda presents the most challenging scenario with the arbitrary types and numbers of distractors that only appear in specific views, making it the most difficult dataset, followed by Android and Statue. The details of the datasets are provided in \cref{appendix:dataset}.

\vspace{-0.1in}
\paragraph{Baselines} We compare our method against several state-of-the-art methods for robust learning, including NeRF-W~\citep{nerfw}, NSFF~\citep{nsff}, NeuralDiff~\citep{neuraldiff}, D$^2$NeRF~\citep{d2nerf}, 
and MipNeRF 360~\citep{mip360} with \(l_1\), \(l_2\) and Charbonnier loss following the previous studies~\citep{robustnerf}. Additionally, we compare with RobustNeRF~\citep{robustnerf}, a recent method that handles distractors as outliers in an optimization problem. Notably, our target problem addresses scenarios where temporal continuity is not guaranteed, meaning the types, numbers, and locations of distractors vary for each view, unlike methods like D$^2$NeRF that assume temporal continuity. 

\vspace{-0.1in}
\paragraph{Implementation Details} 
We use Mip-NeRF360~\citep{mip360} with the same network architecture as in RobustNeRF. To enhance training stability, we utilize the Charbonnier loss~\citep{charbonnier1994two}. More details of the implementation are provided in \cref{appendix:implementation}.

\subsection{Main Results}
\label{subsec:results}
We evaluate the quality of synthesized views using PSNR~\citep{dosselmann2005psnr} to measure the peak signal-to-noise ratio, SSIM~\citep{ssim} for structural similarity, and MS-SSIM~\citep{wang2003multiscale} form multi-scale structural similarity, following the convention.

\vspace{-0.1in}
\paragraph{Synthetic Scenes}
In \cref{table:synthetic}, PruNeRF consistently demonstrates superior performance compared with the baselines across all synthetic datasets. On average, our method achieves an improvement of 1.08 in PSNR and 0.002 in MS-SSIM compared with RobustNeRF. These results highlight PruNeRF's effectiveness in accurately identifying and excluding distractors from the trainset, making it a more robust method than the baselines .

\vspace{-0.1in}
\paragraph{Natural Scenes}
As shown in \cref{table:natural}, PruNeRF outperforms the baselines in natural scenes, with an average improvement of 0.8 PSNR over the state-of-the-art method. Notably, our method exhibits more pronounced improvements in challenging datasets such as BabyYoda and Android. To further demonstrate PruNeRF's effectiveness, we present qualitative results in \cref{fig:qual_eval1} and \cref{fig:qual_eval_d2}, highlighting that PruNeRF shows stable synthesis quality in hard-to-learn regions where RobustNeRF struggles. This discrepancy arises because RobustNeRF masks out large, interconnected areas with high loss magnitudes, leading to the unintended exclusion of hard-to-learn regions. In contrast, our method leverages Influence Functions and 3D spatial consistency to accurately discern distractors from hard-to-learn regions.

\renewcommand{\arraystretch}{1.3}
\begin{table}[t]
\centering
\caption{Quantitative evaluation on the natural scenes.}
\setlength{\tabcolsep}{1pt}
\resizebox{1.0\linewidth}{!}{
    \begin{tabular}{lcccccc}
    \toprule
    \multirow{2}{*}{Method} & \multicolumn{2}{c}{Statue} & \multicolumn{2}{c}{Android} & \multicolumn{2}{c}{BabyYoda} \\
    \cline{2-3} \cline{4-5} \cline{6-7}   
    & SSIM $\uparrow$ & PSNR $\uparrow$ & SSIM $\uparrow$ & PSNR $\uparrow$ & SSIM $\uparrow$ & PSNR \\
    \midrule
    Clean & .80 & 23.57 & .71 & 23.10 & .84 & 32.63 \\
    \midrule
    mip-NeRF360 \small{(L$_2$)} & .66 & 19.09 & .65 & 19.35 & .75 & 22.97 \\
    mip-NeRF360 \small{(L$_1$)} & .72 & 19.55 & .66 & 19.38 & .75 & 26.15 \\
    mip-NeRF360 \small{(Ch.)} & .73 & 19.64 & .66 & 19.53 & .80 & 25.22 \\
    D$^2$NeRF & .49 & 19.09 & .57 & 20.61 & .65 & 17.32 \\
    RobustNeRF & .75 & 20.89 & .65 & 21.72 & .83 & 30.87 \\
    PruNeRF & \textbf{.76} & \textbf{21.31} & \textbf{.72} & \textbf{22.86} & \textbf{.83} & \textbf{31.73}\\
    \bottomrule
    \label{table:natural}
    \end{tabular}
    }
\vspace{-0.2in}
\end{table}

\begin{figure*}[t]
    \begin{center}
    \begin{subfigure}[t]{.24\textwidth}
    \includegraphics[width=1.\textwidth]{{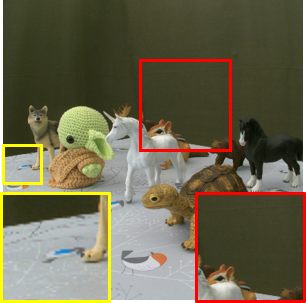}}
    \caption{Ground Truth}
    \end{subfigure}
    \begin{subfigure}[t]{.24\textwidth}
    \includegraphics[width=1.\textwidth]{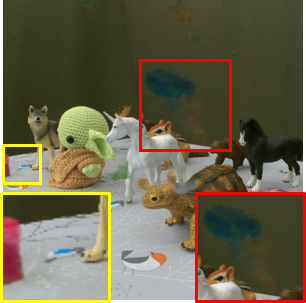}
    \caption{IF}
    \end{subfigure}
    \begin{subfigure}[t]{.24\textwidth}
    \includegraphics[width=1.\textwidth]{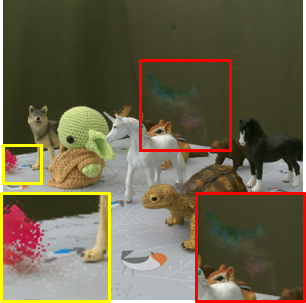}
    \caption{IF \& 3D Cons.}
    \end{subfigure}
    \begin{subfigure}[t]{.24\textwidth}
    \includegraphics[width=1.\textwidth]{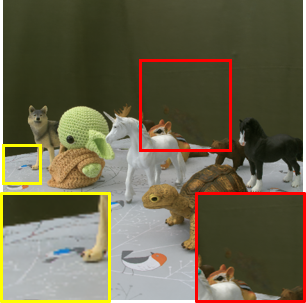}
    \caption{PruNeRF}
    \end{subfigure}
    \end{center}
    \vspace{-0.1in}
    \caption{Qualitative results of ablation study. Using Influence Function alone misses some distractor pixels, but integrating 3D spatial consistency and applying pixel-to-segment refinement produce more accurate results.}
    \label{fig:ablation}
    \vspace{-0.2in}
\end{figure*}

\subsection{Ablation Study}
\label{subsec:ablation}
To demonstrate the contributions of each component including Influence Functions, 3D spatial consistency, and pixel-to-segment refinement, we present ablation studies on natural scene datasets: Statue, Android, and BabyYoda. When evaluating the contribution of Influence Functions, denoted as '+IF' in \cref{table:ablation}, we employ the Otsu algorithm~\citep{otsu} to determine the threshold of Self-Influence scores for pruning. This well-known method divides data into two distributions by maximizing inter-class variance, thereby reducing the reliance on human intervention in threshold selection. As shown in \cref{table:ablation}, while incorporating Influence Functions alone leads to performance improvement, achieving more accurate results necessitates considering 3D spatial consistency and applying pixel-to-segment refinement. Additionally, the qualitative results in \cref{fig:ablation} also show that adding these components progressively leads to clearer removal of distractors. 

Additionally, we conducted an ablation study on the base distraction measurements, maintaining 3D spatial consistency and pixel-to-segment refinement while evaluating loss, gradient-norm, and Influence Function. The results indicate that the Influence Function performs better than loss and gradient-norm, with detailed findings provided in \cref{appen:ablation}. As mentioned earlier, our method's notable advantage is its compatibility with various metrics for measuring sample-wise distraction in a plug-and-play manner.

\begin{table}[t]
\caption{Quantitative results of ablation study (PSNR).}
\centering
\small
\resizebox{1.0\columnwidth}{!}{%
\begin{tabular}{lccc}
\toprule
{Method} & Statue & Android & BabyYoda \\
\midrule
mip-NeRF360 \small{(Ch.)} & 19.64 & 19.53 & 25.22 \\
\hspace{0.5em} + IF & 19.90 & 20.23 & 30.40 \\
\hspace{0.5em} + IF \& 3D Cons. & 20.93 & 20.93 & 31.32 \\
\hspace{0.5em} + PruNeRF & \textbf{21.31} & \textbf{21.22} & \textbf{31.73} \\
\bottomrule
\end{tabular}
}
\label{table:ablation}
\end{table}

\section{Related Work}
\paragraph{Robust Learning in NeRF}
There are several works focusing on NeRF's reliance on well-curated training data. NeRF-W~\citep{nerfw} facilitates training with unstructured internet images by using uncertainty to distinguish static from transient objects, though it has performance limitations. Other studies~\citep{d2nerf, nsff, dynamic, st-nerf} separate static and dynamic components using temporally ordered training images, posing significant challenges. Block-NeRF~\citep{block} employs a semantic segmentation model to identify and exclude undesirable objects, but it is not reliable for objects not present in the pre-training stage. RobustNeRF~\citep{robustnerf} assumes transient objects occupy large, connected regions with high loss values and uses trimmed least squares to train NeRF. However, it struggles to discern hard-to-learn regions from distractor regions, leading to potential artifacts.

\vspace{-0.1in}
\paragraph{Influence Functions}
Influence Function~\citep{inf_1974, inf_blackbox}) and its approximations \citep{tracin, schioppa2022arnoldi} have been utilized in various deep learning tasks by estimating the influence of a data point on the model's prediction. Also, Self-Influence is often used when detecting minority samples, such as mislabeled ones, without a validation set~\citep{resolving_noisy_label, udsif}.

\vspace{-0.1in}
\paragraph{Segmentation}
Image segmentation plays an important role in visual understanding by dividing images into distinct semantic parts.
Recently, several studies~\citep{ding2022decoupling, sam, zou2023segment} have demonstrated remarkable zero-shot segmentation performance. Among these studies, SAM~\citep{sam} has garnered significant attention for its outstanding zero-shot segmentation performance across various datasets for panoptic segmentation supporting prompt input by a user.

\section{Conclusion}
In this paper, we proposed PruNeRF, a segment-centric dataset pruning framework for Neural Radiance Fields (NeRF) that leverages 3D spatial consistency to effectively identify and exclude distractors based on Influence Function scores. Furthermore, we integrate pixel-to-segment refinement via zero-shot segmentation for accurate identification. PruNeRF alleviates the difficulties of dataset construction in NeRF by pruning unintended objects, making it a practical solution in real-world scenarios. Future directions can be optimizing the entire framework in an end-to-end manner or handling scenarios with sparse views, further enhancing the robustness and applicability.

\clearpage
\section*{Impact Statement}
This paper aims to alleviate the difficulties of dataset construction for Neural Radiance Fields (NeRF). We foresee no immediate ethical concerns or negative societal consequences from this work.

\section*{Acknowledgements}
This work was partly supported by Institute of Information \& communications Technology Planning \& Evaluation (IITP) grants (No.2022-0-00984, No.2019-0-00075, Artificial Intelligence Graduate School Program (KAIST)), National Research Foundation of Korea (NRF) grants (RS-2023-00209060, A Study on Optimization and Network Interpretation Method for Large-Scale Machine Learning) funded by the Korea government (MSIT), and KAIST-NAVER Hypercreative AI Center.

\bibliography{main}

\begin{thebibliography}{35}
\providecommand{\natexlab}[1]{#1}
\providecommand{\url}[1]{\texttt{#1}}
\expandafter\ifx\csname urlstyle\endcsname\relax
  \providecommand{\doi}[1]{doi: #1}\else
  \providecommand{\doi}{doi: \begingroup \urlstyle{rm}\Url}\fi

\bibitem[Arpit et~al.(2017)Arpit, Jastrz{k{e}}bski, Ballas, Krueger, Bengio, Kanwal, Maharaj, Fischer, Courville, Bengio, and Lacoste-Julien]{memorization2}
Arpit, D., Jastrz{k{e}}bski, S., Ballas, N., Krueger, D., Bengio, E., Kanwal, M.~S., Maharaj, T., Fischer, A., Courville, A., Bengio, Y., and Lacoste-Julien, S.
\newblock A closer look at memorization in deep networks.
\newblock In Precup, D. and Teh, Y.~W. (eds.), \emph{Proceedings of the 34th International Conference on Machine Learning}, volume~70 of \emph{Proceedings of Machine Learning Research}, pp.\  233--242. PMLR, 06--11 Aug 2017.
\newblock URL \url{https://proceedings.mlr.press/v70/arpit17a.html}.

\bibitem[Barron et~al.(2022)Barron, Mildenhall, Verbin, Srinivasan, and Hedman]{mip360}
Barron, J.~T., Mildenhall, B., Verbin, D., Srinivasan, P.~P., and Hedman, P.
\newblock Mip-nerf 360: Unbounded anti-aliased neural radiance fields.
\newblock In \emph{Proceedings of the IEEE/CVF Conference on Computer Vision and Pattern Recognition}, pp.\  5470--5479, 2022.

\bibitem[Charbonnier et~al.(1994)Charbonnier, Blanc-Feraud, Aubert, and Barlaud]{charbonnier1994two}
Charbonnier, P., Blanc-Feraud, L., Aubert, G., and Barlaud, M.
\newblock Two deterministic half-quadratic regularization algorithms for computed imaging.
\newblock In \emph{Proceedings of 1st international conference on image processing}, volume~2, pp.\  168--172. IEEE, 1994.

\bibitem[Ding et~al.(2022)Ding, Xue, Xia, and Dai]{ding2022decoupling}
Ding, J., Xue, N., Xia, G.-S., and Dai, D.
\newblock Decoupling zero-shot semantic segmentation.
\newblock In \emph{Proceedings of the IEEE/CVF Conference on Computer Vision and Pattern Recognition}, pp.\  11583--11592, 2022.

\bibitem[Dosselmann \& Yang(2005)Dosselmann and Yang]{dosselmann2005psnr}
Dosselmann, R. and Yang, X.~D.
\newblock Existing and emerging image quality metrics.
\newblock In \emph{Canadian Conference on Electrical and Computer Engineering, 2005.}, pp.\  1906--1913, 2005.

\bibitem[Gao et~al.(2021)Gao, Saraf, Kopf, and Huang]{dynamic}
Gao, C., Saraf, A., Kopf, J., and Huang, J.-B.
\newblock Dynamic view synthesis from dynamic monocular video.
\newblock In \emph{Proceedings of the IEEE/CVF International Conference on Computer Vision}, pp.\  5712--5721, 2021.

\bibitem[Hampel(1974)]{inf_1974}
Hampel, F.~R.
\newblock The influence curve and its role in robust estimation.
\newblock \emph{Journal of the american statistical association}, 69\penalty0 (346):\penalty0 383--393, 1974.

\bibitem[Huang et~al.(2021)Huang, Geng, and Li]{gradnorm}
Huang, R., Geng, A., and Li, Y.
\newblock On the importance of gradients for detecting distributional shifts in the wild.
\newblock \emph{Advances in Neural Information Processing Systems}, 34:\penalty0 677--689, 2021.

\bibitem[Killamsetty et~al.(2021)Killamsetty, Zhao, Chen, and Iyer]{gradnorm2}
Killamsetty, K., Zhao, X., Chen, F., and Iyer, R.
\newblock Retrieve: Coreset selection for efficient and robust semi-supervised learning.
\newblock \emph{Advances in Neural Information Processing Systems}, 34:\penalty0 14488--14501, 2021.

\bibitem[Kingma \& Ba(2014)Kingma and Ba]{kingma2014adam}
Kingma, D.~P. and Ba, J.
\newblock Adam: A method for stochastic optimization.
\newblock \emph{arXiv preprint arXiv:1412.6980}, 2014.

\bibitem[Kirillov et~al.(2019)Kirillov, He, Girshick, Rother, and Doll{\'a}r]{kirillov2019panoptic}
Kirillov, A., He, K., Girshick, R., Rother, C., and Doll{\'a}r, P.
\newblock Panoptic segmentation.
\newblock In \emph{Proceedings of the IEEE/CVF conference on computer vision and pattern recognition}, pp.\  9404--9413, 2019.

\bibitem[Kirillov et~al.(2023)Kirillov, Mintun, Ravi, Mao, Rolland, Gustafson, Xiao, Whitehead, Berg, Lo, et~al.]{sam}
Kirillov, A., Mintun, E., Ravi, N., Mao, H., Rolland, C., Gustafson, L., Xiao, T., Whitehead, S., Berg, A.~C., Lo, W.-Y., et~al.
\newblock Segment anything.
\newblock \emph{arXiv preprint arXiv:2304.02643}, 2023.

\bibitem[Koh \& Liang(2017)Koh and Liang]{inf_blackbox}
Koh, P.~W. and Liang, P.
\newblock Understanding black-box predictions via influence functions.
\newblock In \emph{International conference on machine learning}, pp.\  1885--1894. PMLR, 2017.

\bibitem[Kong et~al.(2022)Kong, Shen, and Huang]{resolving_noisy_label}
Kong, S., Shen, Y., and Huang, L.
\newblock Resolving training biases via influence-based data relabeling.
\newblock In \emph{ICLR}, 2022.
\newblock URL \url{https://openreview.net/forum?id=EskfH0bwNVn}.

\bibitem[Kosiorek et~al.(2021)Kosiorek, Strathmann, Zoran, Moreno, Schneider, Mokr{\'a}, and Rezende]{nerfvae}
Kosiorek, A.~R., Strathmann, H., Zoran, D., Moreno, P., Schneider, R., Mokr{\'a}, S., and Rezende, D.~J.
\newblock Nerf-vae: A geometry aware 3d scene generative model.
\newblock In \emph{International Conference on Machine Learning}, pp.\  5742--5752. PMLR, 2021.

\bibitem[Lee et~al.(2018)Lee, Lee, Lee, and Shin]{mahalanobis}
Lee, K., Lee, K., Lee, H., and Shin, J.
\newblock A simple unified framework for detecting out-of-distribution samples and adversarial attacks.
\newblock \emph{Advances in neural information processing systems}, 31, 2018.

\bibitem[Li et~al.(2021)Li, Niklaus, Snavely, and Wang]{nsff}
Li, Z., Niklaus, S., Snavely, N., and Wang, O.
\newblock Neural scene flow fields for space-time view synthesis of dynamic scenes.
\newblock In \emph{Proceedings of the IEEE/CVF Conference on Computer Vision and Pattern Recognition}, pp.\  6498--6508, 2021.

\bibitem[Liang et~al.(2018)Liang, Li, and Srikant]{odin}
Liang, S., Li, Y., and Srikant, R.
\newblock Enhancing the reliability of out-of-distribution image detection in neural networks.
\newblock In \emph{International Conference on Learning Representations}, 2018.

\bibitem[Martin-Brualla et~al.(2021)Martin-Brualla, Radwan, Sajjadi, Barron, Dosovitskiy, and Duckworth]{nerfw}
Martin-Brualla, R., Radwan, N., Sajjadi, M.~S., Barron, J.~T., Dosovitskiy, A., and Duckworth, D.
\newblock Nerf in the wild: Neural radiance fields for unconstrained photo collections.
\newblock In \emph{Proceedings of the IEEE/CVF Conference on Computer Vision and Pattern Recognition}, pp.\  7210--7219, 2021.

\bibitem[Mildenhall et~al.(2021)Mildenhall, Srinivasan, Tancik, Barron, Ramamoorthi, and Ng]{nerf}
Mildenhall, B., Srinivasan, P.~P., Tancik, M., Barron, J.~T., Ramamoorthi, R., and Ng, R.
\newblock Nerf: Representing scenes as neural radiance fields for view synthesis.
\newblock \emph{Communications of the ACM}, 65\penalty0 (1):\penalty0 99--106, 2021.

\bibitem[Otsu(1979)]{otsu}
Otsu, N.
\newblock A threshold selection method from gray-level histograms.
\newblock \emph{IEEE transactions on systems, man, and cybernetics}, 9\penalty0 (1):\penalty0 62--66, 1979.

\bibitem[Pruthi et~al.(2020)Pruthi, Liu, Kale, and Sundararajan]{tracin}
Pruthi, G., Liu, F., Kale, S., and Sundararajan, M.
\newblock Estimating training data influence by tracing gradient descent.
\newblock \emph{Advances in Neural Information Processing Systems}, 33:\penalty0 19920--19930, 2020.

\bibitem[Sabour et~al.(2023)Sabour, Vora, Duckworth, Krasin, Fleet, and Tagliasacchi]{robustnerf}
Sabour, S., Vora, S., Duckworth, D., Krasin, I., Fleet, D.~J., and Tagliasacchi, A.
\newblock Robustnerf: Ignoring distractors with robust losses.
\newblock In \emph{Proceedings of the IEEE/CVF Conference on Computer Vision and Pattern Recognition}, pp.\  20626--20636, 2023.

\bibitem[Schioppa et~al.(2022{\natexlab{a}})Schioppa, Zablotskaia, Vilar, and Sokolov]{arnoldi}
Schioppa, A., Zablotskaia, P., Vilar, D., and Sokolov, A.
\newblock Scaling up influence functions.
\newblock In \emph{Proceedings of the AAAI Conference on Artificial Intelligence}, volume~36, pp.\  8179--8186, 2022{\natexlab{a}}.

\bibitem[Schioppa et~al.(2022{\natexlab{b}})Schioppa, Zablotskaia, Vilar, and Sokolov]{schioppa2022arnoldi}
Schioppa, A., Zablotskaia, P., Vilar, D., and Sokolov, A.
\newblock Scaling up influence functions.
\newblock In \emph{Proceedings of the AAAI Conference on Artificial Intelligence}, volume~36, pp.\  8179--8186, 2022{\natexlab{b}}.

\bibitem[Selvaraju et~al.(2017)Selvaraju, Cogswell, Das, Vedantam, Parikh, and Batra]{gradcam}
Selvaraju, R.~R., Cogswell, M., Das, A., Vedantam, R., Parikh, D., and Batra, D.
\newblock Grad-cam: Visual explanations from deep networks via gradient-based localization.
\newblock In \emph{Proceedings of the IEEE International Conference on Computer Vision (ICCV)}, Oct 2017.

\bibitem[Tancik et~al.(2022)Tancik, Casser, Yan, Pradhan, Mildenhall, Srinivasan, Barron, and Kretzschmar]{block}
Tancik, M., Casser, V., Yan, X., Pradhan, S., Mildenhall, B., Srinivasan, P.~P., Barron, J.~T., and Kretzschmar, H.
\newblock Block-nerf: Scalable large scene neural view synthesis.
\newblock In \emph{Proceedings of the IEEE/CVF Conference on Computer Vision and Pattern Recognition}, pp.\  8248--8258, 2022.

\bibitem[Tschernezki et~al.(2021)Tschernezki, Larlus, and Vedaldi]{neuraldiff}
Tschernezki, V., Larlus, D., and Vedaldi, A.
\newblock Neuraldiff: Segmenting 3d objects that move in egocentric videos.
\newblock In \emph{2021 International Conference on 3D Vision (3DV)}, pp.\  910--919. IEEE, 2021.

\bibitem[Wang et~al.(2003)Wang, Simoncelli, and Bovik]{wang2003multiscale}
Wang, Z., Simoncelli, E.~P., and Bovik, A.~C.
\newblock Multiscale structural similarity for image quality assessment.
\newblock In \emph{The Thrity-Seventh Asilomar Conference on Signals, Systems \& Computers, 2003}, volume~2, pp.\  1398--1402. Ieee, 2003.

\bibitem[Wang et~al.(2004)Wang, Bovik, Sheikh, and Simoncelli]{ssim}
Wang, Z., Bovik, A.~C., Sheikh, H.~R., and Simoncelli, E.~P.
\newblock Image quality assessment: from error visibility to structural similarity.
\newblock \emph{IEEE transactions on image processing}, 13\penalty0 (4):\penalty0 600--612, 2004.

\bibitem[Wang et~al.(2020)Wang, Zhu, Dong, He, and Huang]{udsif}
Wang, Z., Zhu, H., Dong, Z., He, X., and Huang, S.
\newblock Less is better: Unweighted data subsampling via influence function.
\newblock In \emph{The Thirty-Fourth {AAAI} Conference on Artificial Intelligence, {AAAI} 2020, The Thirty-Second Innovative Applications of Artificial Intelligence Conference, {IAAI} 2020, The Tenth {AAAI} Symposium on Educational Advances in Artificial Intelligence, {EAAI} 2020, New York, NY, USA, February 7-12, 2020}, pp.\  6340--6347. {AAAI} Press, 2020.

\bibitem[Wu et~al.(2022)Wu, Zhong, Tagliasacchi, Cole, and Oztireli]{d2nerf}
Wu, T., Zhong, F., Tagliasacchi, A., Cole, F., and Oztireli, C.
\newblock D\^{} 2nerf: Self-supervised decoupling of dynamic and static objects from a monocular video.
\newblock \emph{Advances in Neural Information Processing Systems}, 35:\penalty0 32653--32666, 2022.

\bibitem[Zhang et~al.(2017)Zhang, Bengio, Hardt, Recht, and Vinyals]{memorization}
Zhang, C., Bengio, S., Hardt, M., Recht, B., and Vinyals, O.
\newblock Understanding deep learning requires rethinking generalization.
\newblock In \emph{International Conference on Learning Representations}, 2017.
\newblock URL \url{https://openreview.net/forum?id=Sy8gdB9xx}.

\bibitem[Zhang et~al.(2021)Zhang, Liu, Ye, Zhao, Zhang, Wu, Zhang, Yu, and Xu]{st-nerf}
Zhang, J., Liu, X., Ye, X., Zhao, F., Zhang, Y., Wu, M., Zhang, Y., Yu, J., and Xu, L.
\newblock Editable free-viewpoint video using a layered neural representation.
\newblock 2021.

\bibitem[Zou et~al.(2023)Zou, Yang, Zhang, Li, Li, Gao, and Lee]{zou2023segment}
Zou, X., Yang, J., Zhang, H., Li, F., Li, L., Gao, J., and Lee, Y.~J.
\newblock Segment everything everywhere all at once.
\newblock \emph{arXiv preprint arXiv:2304.06718}, 2023.

\end{thebibliography}
\bibliographystyle{icml2024}

\newpage
\appendix
\onecolumn
\section{Datasets}
\label{appendix:dataset}
\begin{figure}[h]
    \begin{center}
    \includegraphics[width=0.20\textwidth]{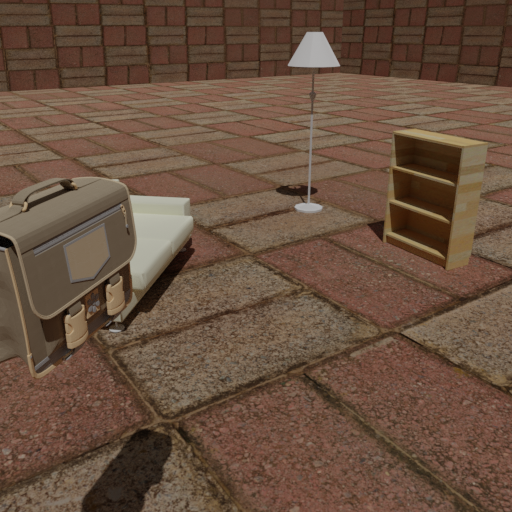}
    \includegraphics[width=0.20\textwidth]{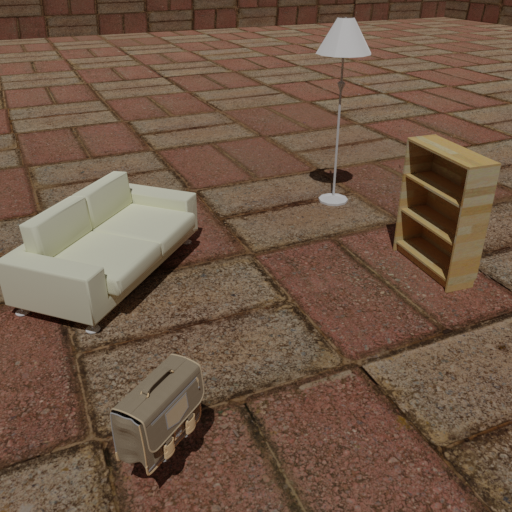}
    \includegraphics[width=0.20\textwidth]{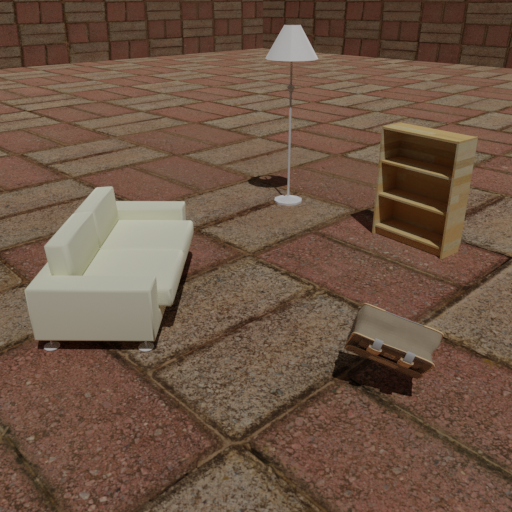}
    \includegraphics[width=0.20\textwidth]{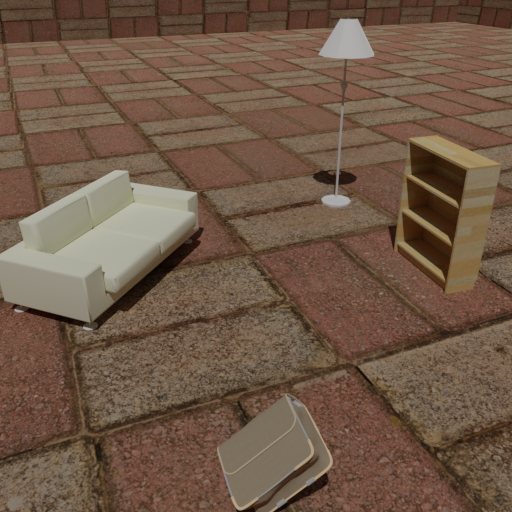}
    \end{center}
    \caption{Example training views from the Kubric bag dataset showing a bag moving as a distractor.}
    \label{fig:kubric}
\end{figure}

The Kubric dataset is a synthetic dataset created with ground-truth masks for moving objects and their shadows~\citep{d2nerf}. It comprises five scenes with one or more dynamic objects from ShapeNet, which can exhibit either rigid or non-rigid motion. The dataset generation process involves moving a virtual camera across 10 keyframes sampled from specified azimuth and altitude ranges, resulting in a 200-frame video sequence used for training. For validation, the camera rotates around the keyframe center, generating 100 views that only display the static background. This setup allows for the creation of masks for dynamic objects and their shadows, providing a detailed means to assess algorithm performance. Including shadows, which are typically missing in existing benchmarks, adds an additional layer of evaluation capability. Each scene comprises 200 training views and 100 test views. We provide examples from the Kubric bag dataset in \cref{fig:kubric}.

\vspace{-0.15in}
\subsection{RobustNeRF} 
\begin{figure}[h]
    \begin{center}
        \includegraphics[width=0.9\textwidth]{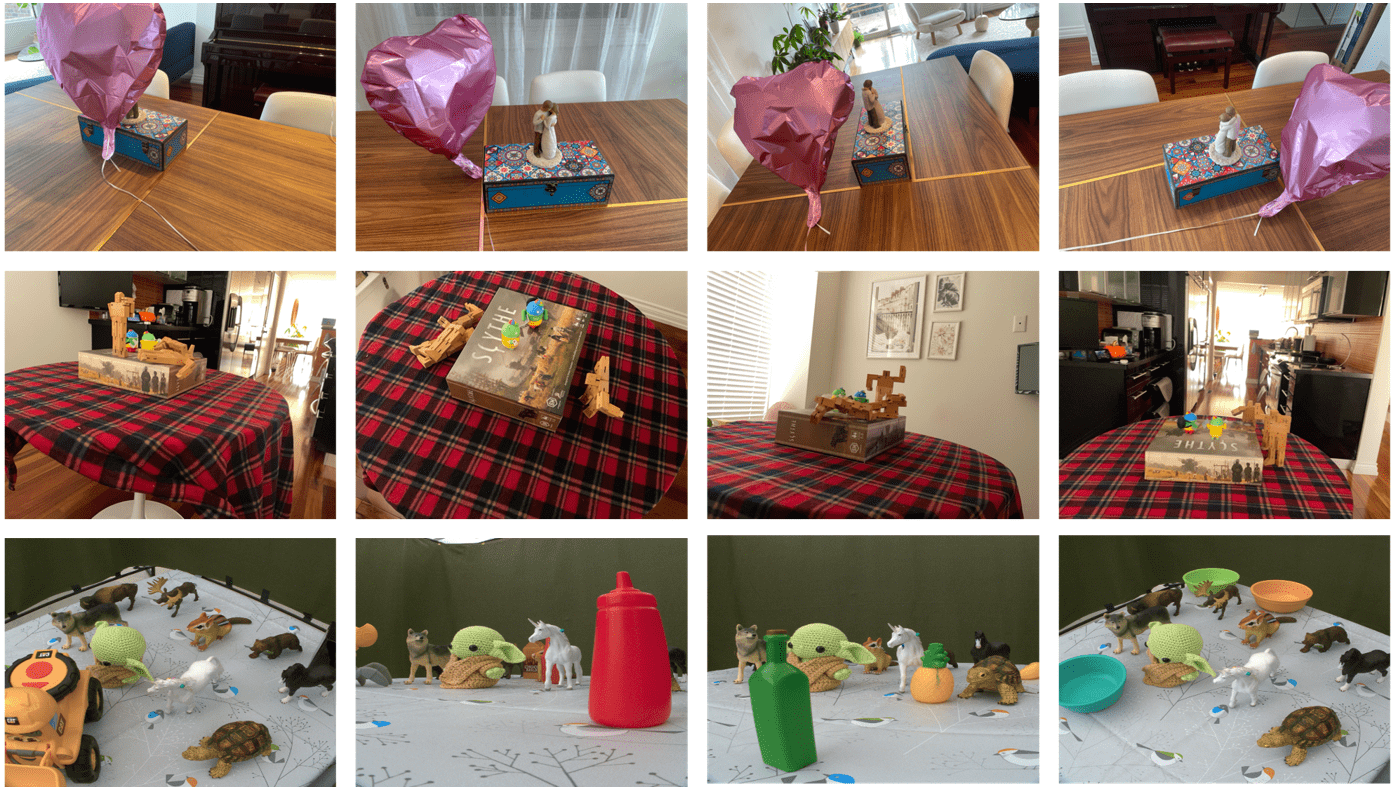}
    \end{center}
    \caption{Examples from the RobustNerf datasets: Statue, Android, and BabyYoda.}
    \label{fig:natural}
\end{figure}

These datasets include several scenes with various distractors, captured in different settings as shown in \cref{fig:natural}. The Statue scene, captured in an apartment, features a small statue on a decorative box with a balloon as a persistent distractor. This scene includes 255 training views and 19 test views. The Android scene, also captured in an apartment, depicts two Android robot figures with three small wooden robots as distractors, changing position in each of the 122 training views, with an additional 10 views for evaluation. The BabyYoda scene, captured in a robotics lab, involves up to 150 unique distractors, with controlled lighting and precise object placement using a robotic arm. This scene contains 109 training views and 202 test views. Camera poses for all scenes were estimated using COLMAP.

\section{Implementation Details}
\label{appendix:implementation}
We use Mip-NeRF360~\citep{mip360} as Proposal MLP with 4 layers and 256 hidden units, NeRF MLP with 8 layers and 1024 hidden units, following the previous work~\citep{robustnerf}. Also, the model is trained for 250k iterations with a batch size of 16,384. We employ the Adam optimizer~\citep{kingma2014adam} with hyperparameters of $\beta_1=0.9$, $\beta_2=0.999$, and $\epsilon=10^{-6}$. The initial learning rate is set to $2\times 10^{-3}$ and exponentially decayed to $2\times10^{-6}$. The first 512 iterations are used for warm-up. To enhance training stability, we utilize the Charbonnier loss~\citep{charbonnier1994two}.

\section{Additional Qualitative Results}
\label{appen:qual_eval}
\begin{figure}[h]
    \begin{center}
    \includegraphics[width=0.24 \textwidth, height=3cm]{{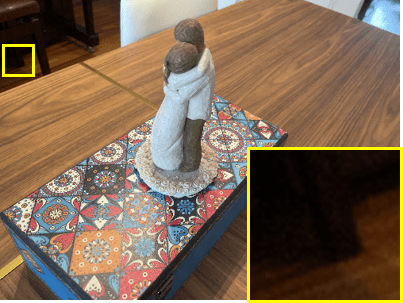}}
    \includegraphics[width=0.24 \textwidth, height=3cm]{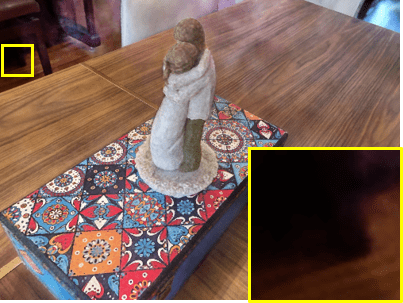}
    \includegraphics[width=0.24 \textwidth, height=3cm]{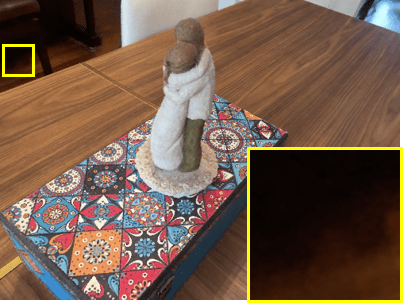}
    \includegraphics[width=0.24 \textwidth, height=3cm]{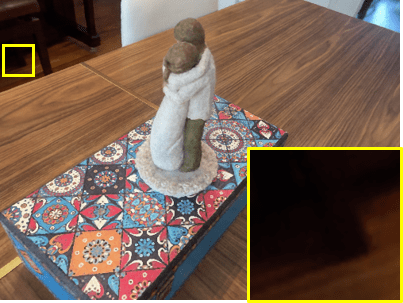}
    \end{center}

    \begin{center}
    \includegraphics[width=0.24 \textwidth, height=3cm]{{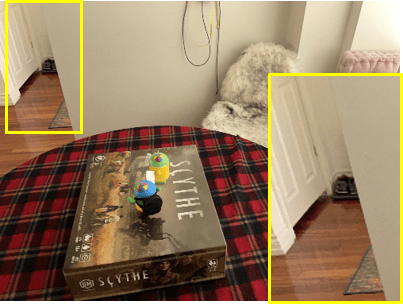}}
    \includegraphics[width=0.24 \textwidth, height=3cm]{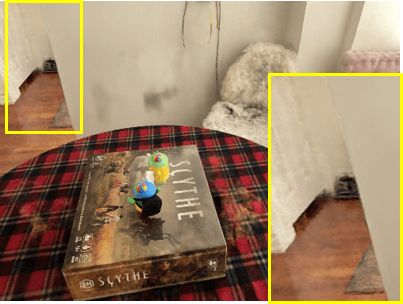}
    \includegraphics[width=0.24 \textwidth, height=3cm]{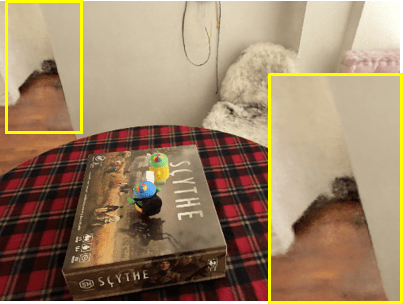}
    \includegraphics[width=0.24 \textwidth, height=3cm]{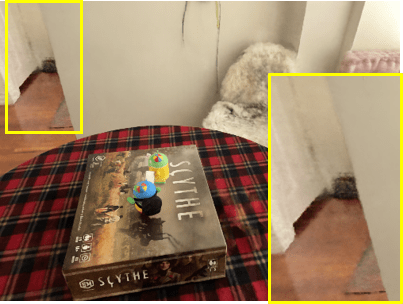}
    \end{center}

    \begin{center}
    \begin{subfigure}[t]{0.24\textwidth}
    \includegraphics[width=1.\textwidth, height=3cm]{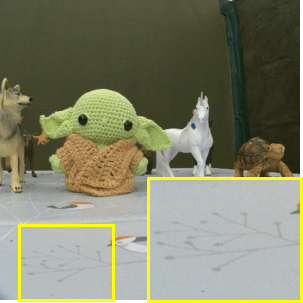}
    \caption{Ground Truth}
    \end{subfigure}
    \begin{subfigure}[t]{0.24\textwidth}
    \includegraphics[width=1.\textwidth, height=3cm]{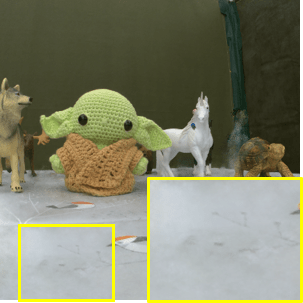}
    \caption{MipNeRF360}
    \end{subfigure}
    \begin{subfigure}[t]{0.24\textwidth}
    \includegraphics[width=1.\textwidth, height=3cm]{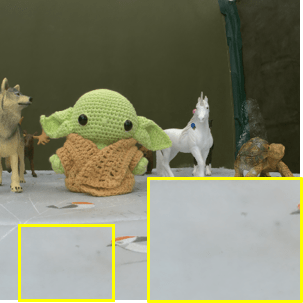}
    \caption{RobustNeRF}
    \end{subfigure}
    \begin{subfigure}[t]{0.24\textwidth}
    \includegraphics[width=1.\textwidth, height=3cm]{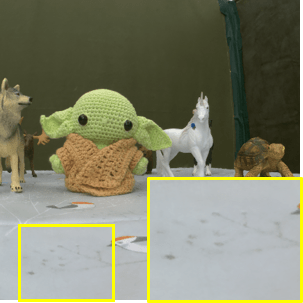}
    \caption{PruNeRF}
    \end{subfigure}
    \end{center}
    \caption{Additional qualitative results on Statue, Android, and BabyYoda.}
    \label{appen:fig:qual_eval}
\end{figure}

In \cref{fig:qual_eval1}, we provided qualitative results on Statue, Android, and BabyYoda. In \cref{appen:fig:qual_eval} we show additional qualitative results on them. 
 
\clearpage
\section{Ablation study on distraction measuring metrics}
\label{appen:ablation}
we conducted an ablation study on the base distraction measurements, maintaining 3D spatial consistency and pixel-to-segment refinement while evaluating loss, gradient-norm, and Influence Function. The results, as detailed in the \cref{table:ablation_merics}, indicate that Influence Function exhibits better performance compared with alternatives. Additionally, to further enhance performance, we expanded our analyses by exploring another approximated version of the Influence Function, thereby enriching our study. In our paper, to mitigate the computational cost of calculating the Hessian inverse in the Influence Function, we adopted a common strategy that limits Hessian computation only to the last layer of the model. While this approach is common in 2D image classification where the last layer typically serves as the classifier, it may lead to inaccuracies due to the different network structures in NeRF. To circumvent this issue, we adopted Arnoldi~\citep{arnoldi}, which is one of the full-model parameter approximation techniques and is well-known for its efficient implementation. The results indicate that using Arnoldi provides further improvement compared with using loss and gradnorm across three natural scenes.

\begin{table}[h]
\caption{Ablation study on distraction measuring metrics (PSNR).}
\centering
\resizebox{0.7\columnwidth}{!}{%
\begin{tabular}{lcccccc}
\toprule
\multirow{2}{*}{Method} & \multicolumn{2}{c}{Statue} & \multicolumn{2}{c}{Android} & \multicolumn{2}{c}{BabyYoda} \\
\cline{2-3} \cline{4-5} \cline{6-7}   
& SSIM $\uparrow$ & PSNR $\uparrow$ & SSIM $\uparrow$ & PSNR $\uparrow$ & SSIM $\uparrow$ & PSNR $\uparrow$\\
\midrule
PruNeRF\hspace{0.2em}\small{$loss$} & .75 & 20.82 & .71 & 22.29 & .82 & 31.14 \\
PruNeRF\hspace{0.2em}\small{$gradnorm$} & .75 & 20.95 & .71 & 22.57 & .82 & 31.47 \\
PruNeRF\hspace{0.2em}\small{$IF_{lastlayer}$} & \underline{.76} & \underline{21.31} & \textbf{.72} & \underline{22.86} & \underline{.83} & \underline{31.73} \\
PruNeRF\hspace{0.2em}\small{$IF_{arnoldi}$} & \textbf{.77} & \textbf{21.46} & \textbf{.72} & \textbf{22.97} & \textbf{.84} & \textbf{31.93} \\
\bottomrule
\end{tabular}
}
\label{table:ablation_merics}
\end{table}

\end{document}